\documentclass[runningheads]{llncs}
\usepackage{graphicx}
\usepackage[table,xcdraw]{xcolor}

\usepackage{tikz}
\usepackage{comment}
\usepackage{amsmath,amssymb} 
\usepackage{color}
\usepackage{subfigure}
\usepackage[colorlinks]{hyperref}
\usepackage{amsfonts}
\usepackage{amsmath}
\usepackage{bm}
\usepackage{multirow}
\usepackage{booktabs}

\usepackage[accsupp]{axessibility}  

\begin{document}
\pagestyle{headings}
\mainmatter
\def\ECCVSubNumber{190}  

\title{Fine-grained Data Distribution Alignment for Post-Training Quantization} 

\titlerunning{Fine-grained Data Distribution Alignment for Post-Training Quantization}
%
\author{Yunshan Zhong$^{1,2}$, Mingbao Lin$^{2,3}$, Mengzhao Chen$^{2}$, Ke Li$^3$, \\ Yunhang Shen$^3$, Fei Chao$^{2}$, Yongjian Wu$^3$, Rongrong Ji$^{1,2}$\thanks{Corresponding Author: rrji@xmu.edu.cn}}
\authorrunning{Yunshan Zhong \emph{et al}.}
%
\institute{$^1$Institute of Artificial Intelligence, Xiamen University \\ $^2$Media Analytics and Computing Lab, Department of Artificial Intelligence, \\School of Informatics, Xiamen University \\ $^3$Tencent Youtu Lab}
\maketitle

\begin{abstract}
While post-training quantization receives popularity mostly due to its evasion in accessing the original complete training dataset, its poor performance also stems from scarce images. To alleviate this limitation, in this paper, we leverage the synthetic data introduced by zero-shot quantization with calibration dataset and propose a fine-grained data distribution alignment (FDDA) method to boost the performance of post-training quantization. The method is based on two important properties of batch normalization statistics (BNS) we observed in deep layers of the trained network, \emph{i.e.}, inter-class separation and intra-class incohesion. To preserve this fine-grained distribution information: 1) We calculate the per-class BNS of the calibration dataset as the BNS centers of each class and propose a BNS-centralized loss to force the synthetic data distributions of different classes to be close to their own centers. 2) We add Gaussian noise into the centers to imitate the incohesion and propose a BNS-distorted loss to force the synthetic data distribution of the same class to be close to the distorted centers. By utilizing these two fine-grained losses, our method manifests the state-of-the-art performance on ImageNet, especially when both the first and last layers are quantized to the low-bit. Code is at \url{https://github.com/zysxmu/FDDA}.
\end{abstract}

\section{Introduction}

Recent years have witnessed the rising of deep neural networks (DNNs) in computer vision. Nevertheless, the increasing model size barricades the deployment of DNNs on resource-limited platforms such as mobile phones, embedding devices, \emph{etc}. To overcome this dilemma, varieties of methods~\cite{han2015learning,whitepaper} are explored to reduce the complexity of DNNs. Network quantization, which represents full-precision DNNs in a low-precision format, emerges as a promising direction~\cite{ACIQ,whitepaper,lin2020rotated}.

\begin{figure}[!t]
\begin{center}
\includegraphics[height=0.4\linewidth]{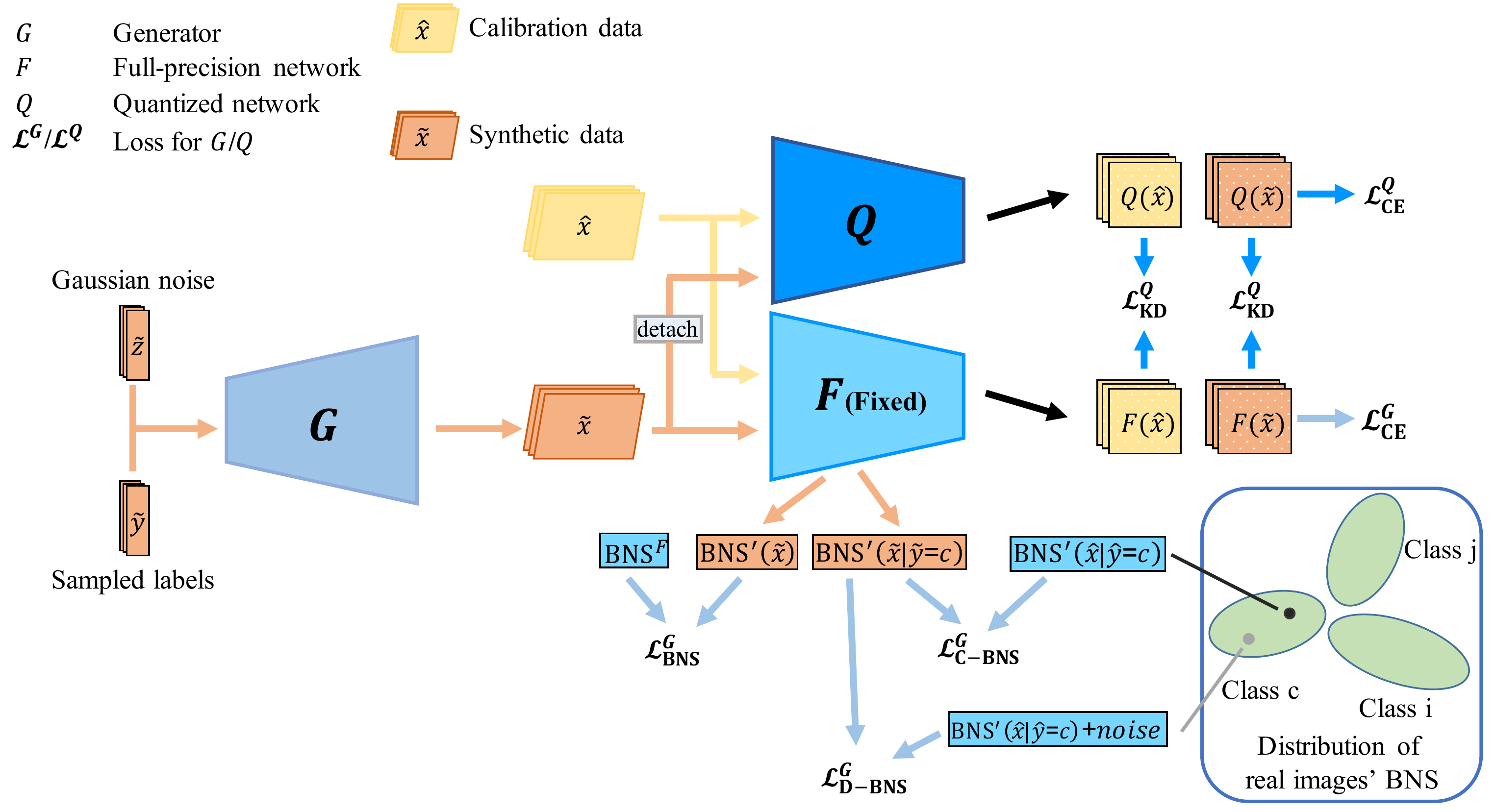}
\end{center}
\vspace{-1em}
\caption{Framework of our FDDA. We deploy a generator $G$ to produce synthetic data supervised by the coarse BNS alignment loss ($\mathcal{L}^G_{\text{BNS}}$), and two proposed fine-grained distribution losses ($\mathcal{L}^G_{\text{C-BNS}}$ and $\mathcal{L}^G_{\text{D-BNS}}$).}
\label{framework}
\vspace{-1.5em}
\end{figure}    

By tuning the quantized DNNs using a small calibration dataset, post-training quantization, a sub-topic of low-precision quantization, has received increasing popularity from both academia and industries. Recent studies manifest that a post-training quantized model in high precision, such as 8-bit, can reach performance on par with its full-precision counterpart~\cite{whitepaper,ACIQ}. However, performance drops severely if being quantized to lower precision such as 4-bit~\cite{li2021brecq}. 
For example, as reported in LAPQ~\cite{LAPQ}, quantizing ResNet-18~\cite{he2016deep} to 8-bit can well retain the accuracy of the full-precision network (around $71.5\%$), but only $60.3\%$ top-1 accuracy can be observed when quantized to 4-bit.
To alleviate this problem, many studies are explored to enhance the low-bit performance. The mainstream can be outlined into two folds.
The first group designs sophisticated quantization methods, such as linear combination of multiple low-bit vectors~\cite{MPwMP}, weight region separation~\cite{fang2020post}, mixed-precision quantization~\cite{MPwMP}, partial quantization~\cite{kryzhanovskiy2021qpp}, \emph{etc}. The second group reformulates the rounding function or loss constraint from an analytical perspective. For example, Nagel \emph{et al}.~\cite{Upordown} derived an adaptive rounding by modeling rounding problem as quadratic constrained binary optimization. By the second-order analysis on rebuilding intermediate outputs, Li \emph{et al}.~\cite{li2021brecq} showed that the best output reconstruction lies in a block unit.

Though great efforts have been made, improvements of these studies are still limited. Besides, the performance gains are usually built on the premise that the first and last layers are quantized to 8-bit~\cite{adaquant,BitSplitStitching}, or even retained in full-precision states~\cite{LAPQ,Upordown}. However, severe performance degradation occurs when all layers are quantized to very low-bit integers (see Table\,\ref{comparison}).
We consider the root cause of significant accuracy degradation in post-training quantization. We attribute it to a lack of training data. Specifically, the low-bit network bears poor representation ability, and a very small calibration dataset cannot support the quantized model to fit well the real data distribution. Many researches on zero-shot quantization are indicated to synthesizing fake images using data optimizer~\cite{yin2020dreaming,zeroq,DSG,Theknowledgewithin} or data generator~\cite{GDFQ,ZAQ,choi2021qimera}. The synthetic data is then used to train the quantized model. Though this manner partly alleviates data lacking, performance drops greatly if simply using synthetic data. For example, Li \emph{et al}.~\cite{li2021brecq} observed only $21.71\%$ top-1 accuracy when quantizing ResNet-18 to 4-bit in a single zero-shot manner~\cite{zeroq}.
So far, combining real calibration dataset with synthetic data remains unexplored in post-training quantization, and we believe, might be a promise of boosting the low-bit performance.

\begin{figure*}[t]
\centering
\subfigure[]{
\includegraphics[height=0.145\linewidth]{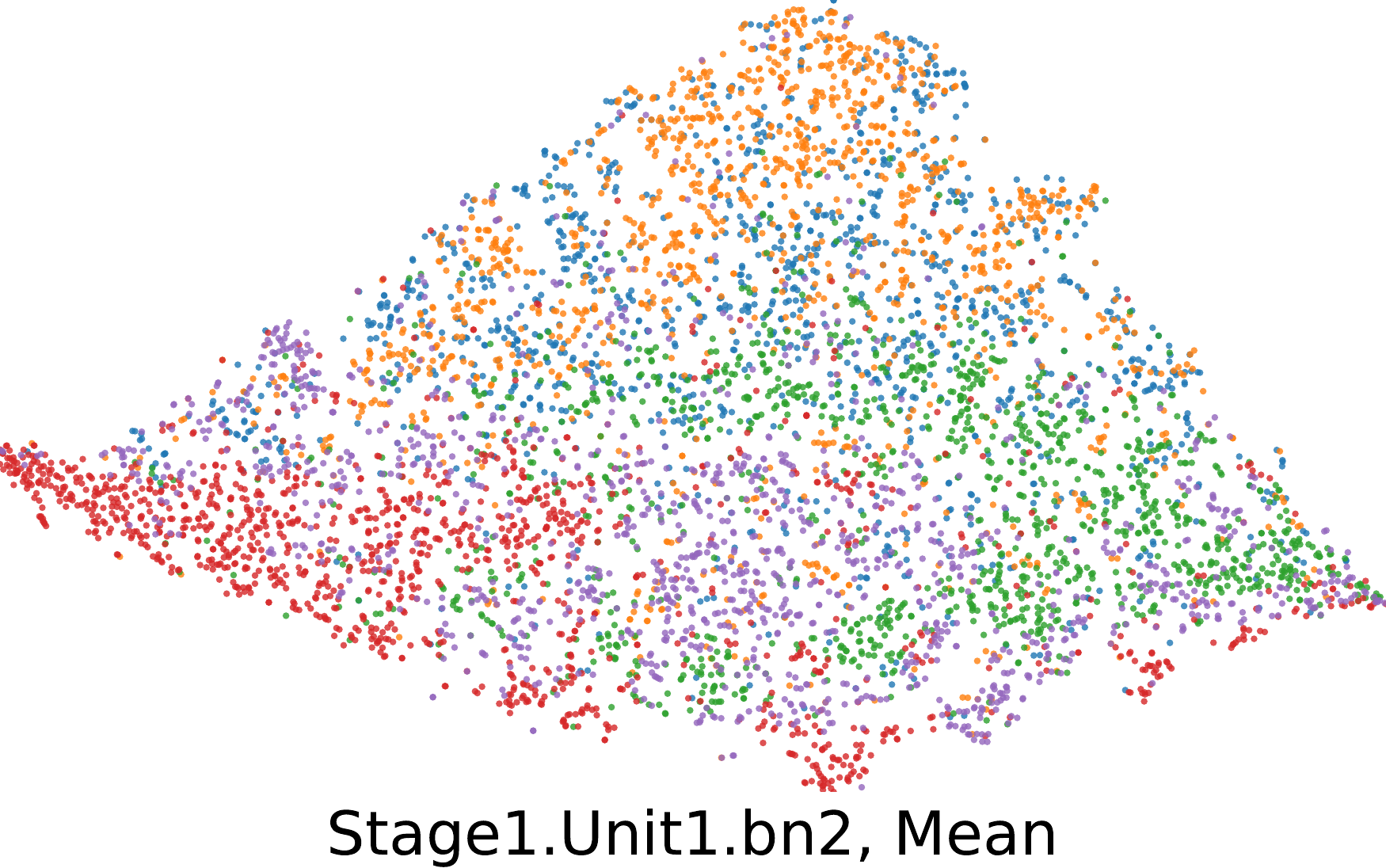} \label{visualization:1}
}
\subfigure[]{
\includegraphics[height=0.145\linewidth]{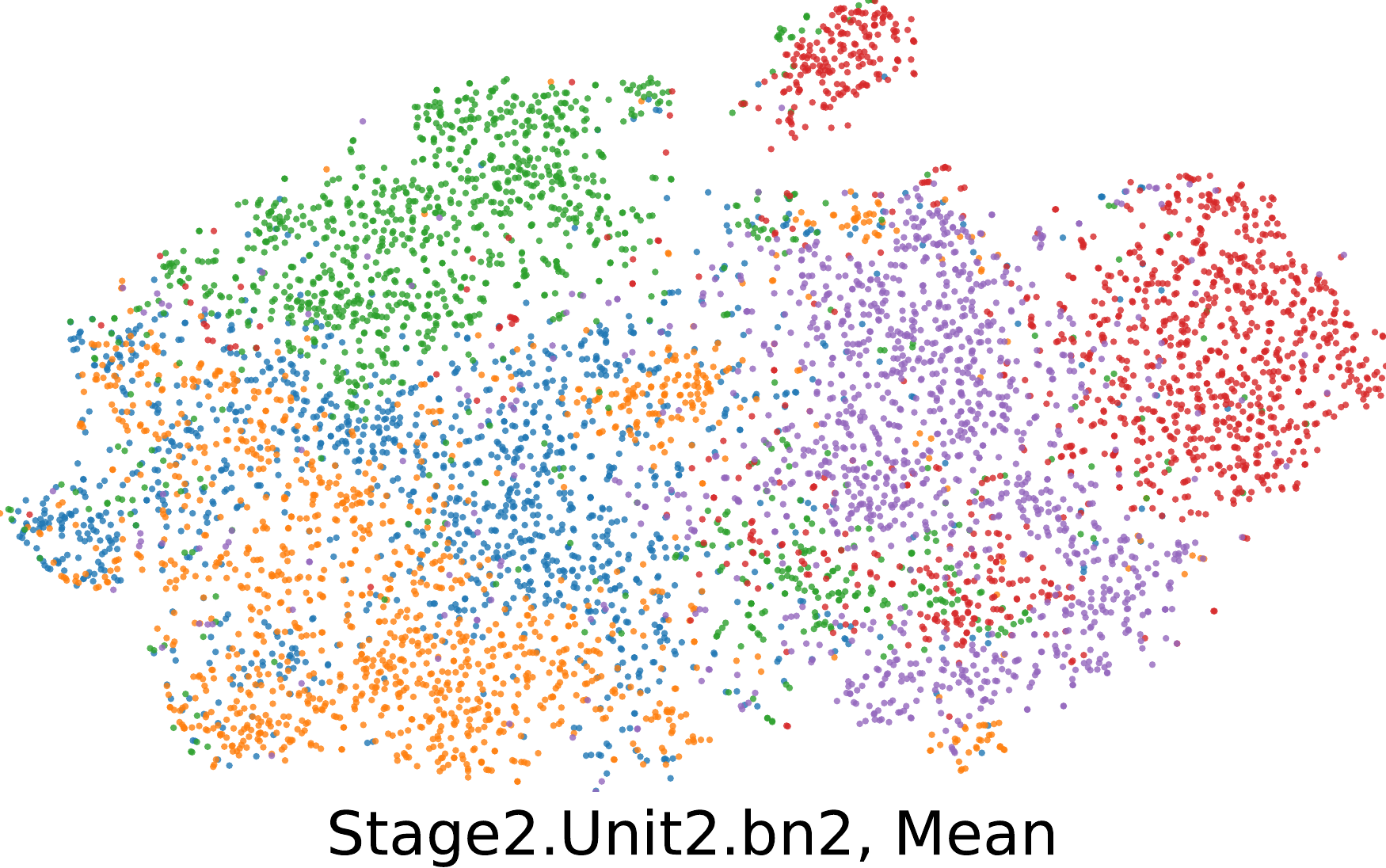} \label{visualization:2} 
}
\subfigure[]{
\includegraphics[height=0.145\linewidth]{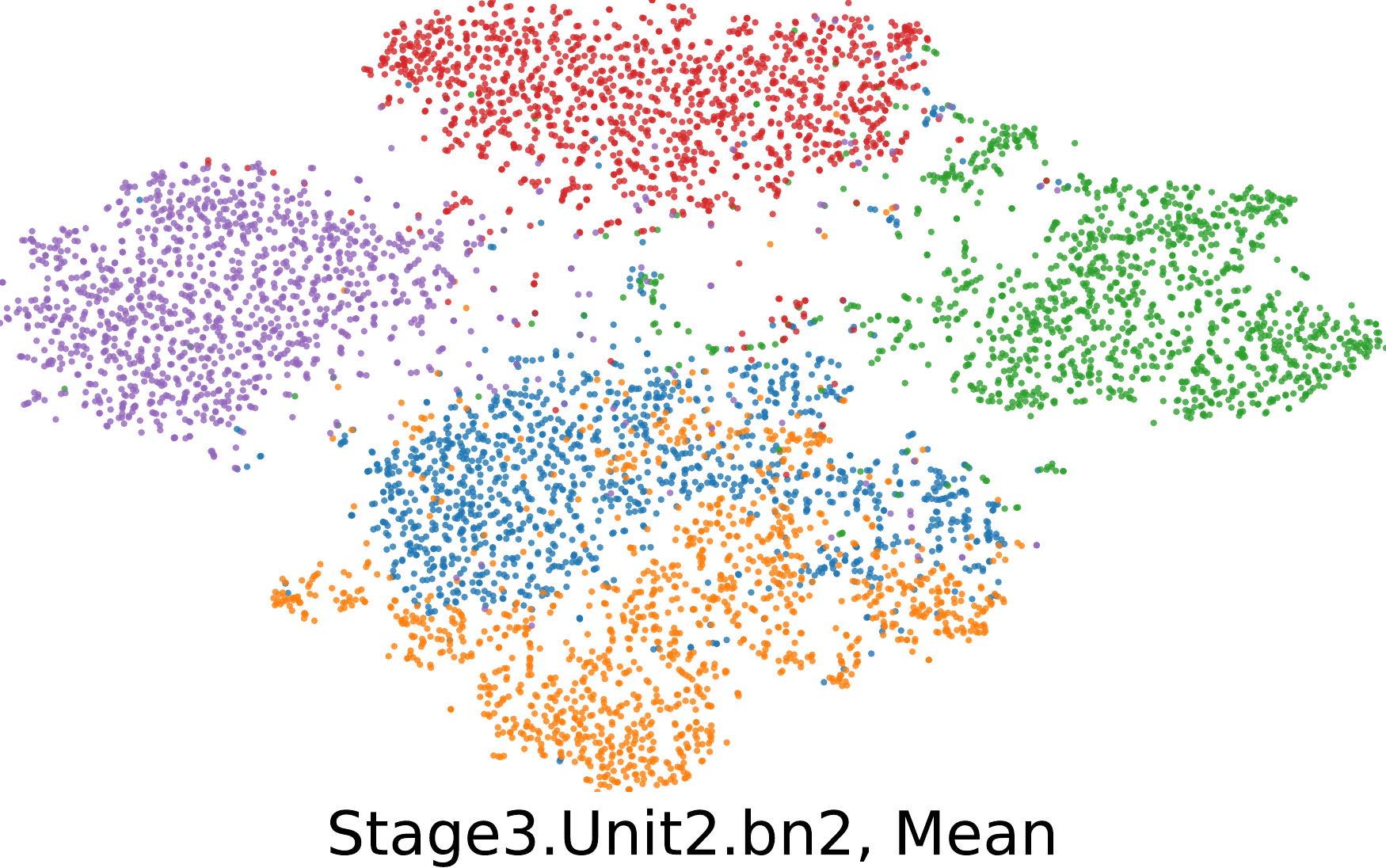} \label{visualization:3} 
}
\subfigure[]{
\includegraphics[height=0.145\linewidth]{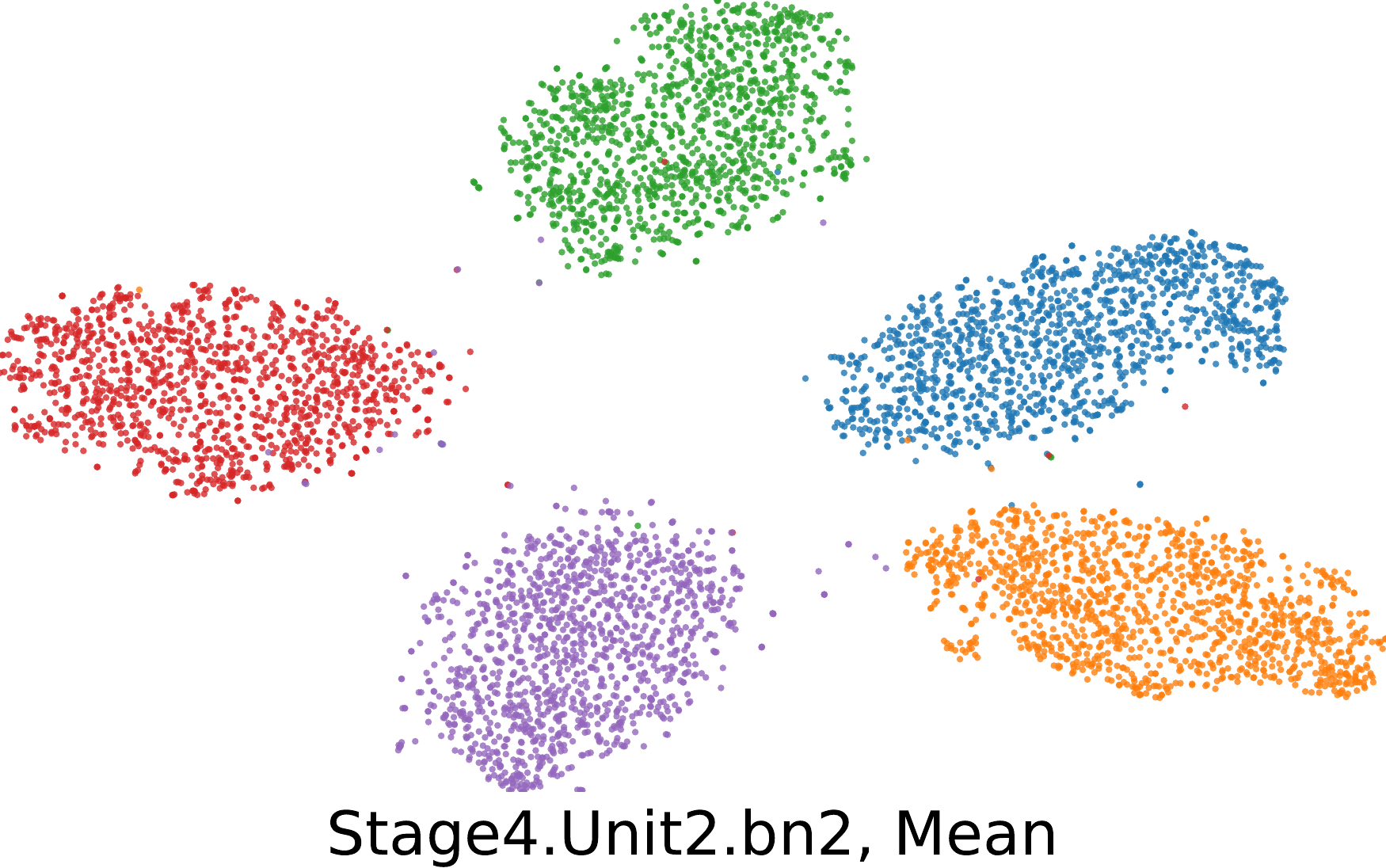}\label{visualization:4}
}

\subfigure[]{
\includegraphics[height=0.145\linewidth]{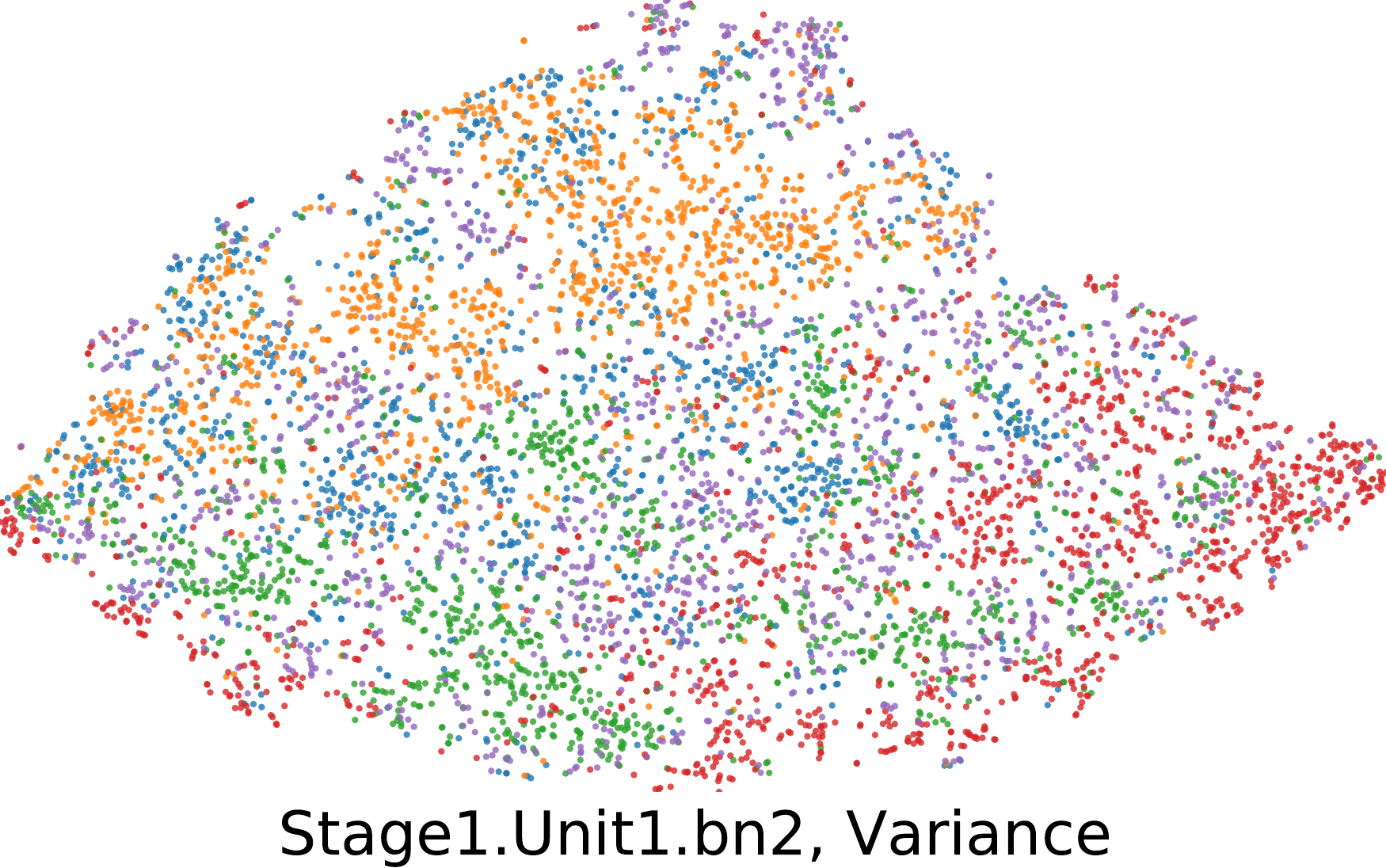} \label{visualization:5}
}
\subfigure[]{
\includegraphics[height=0.145\linewidth]{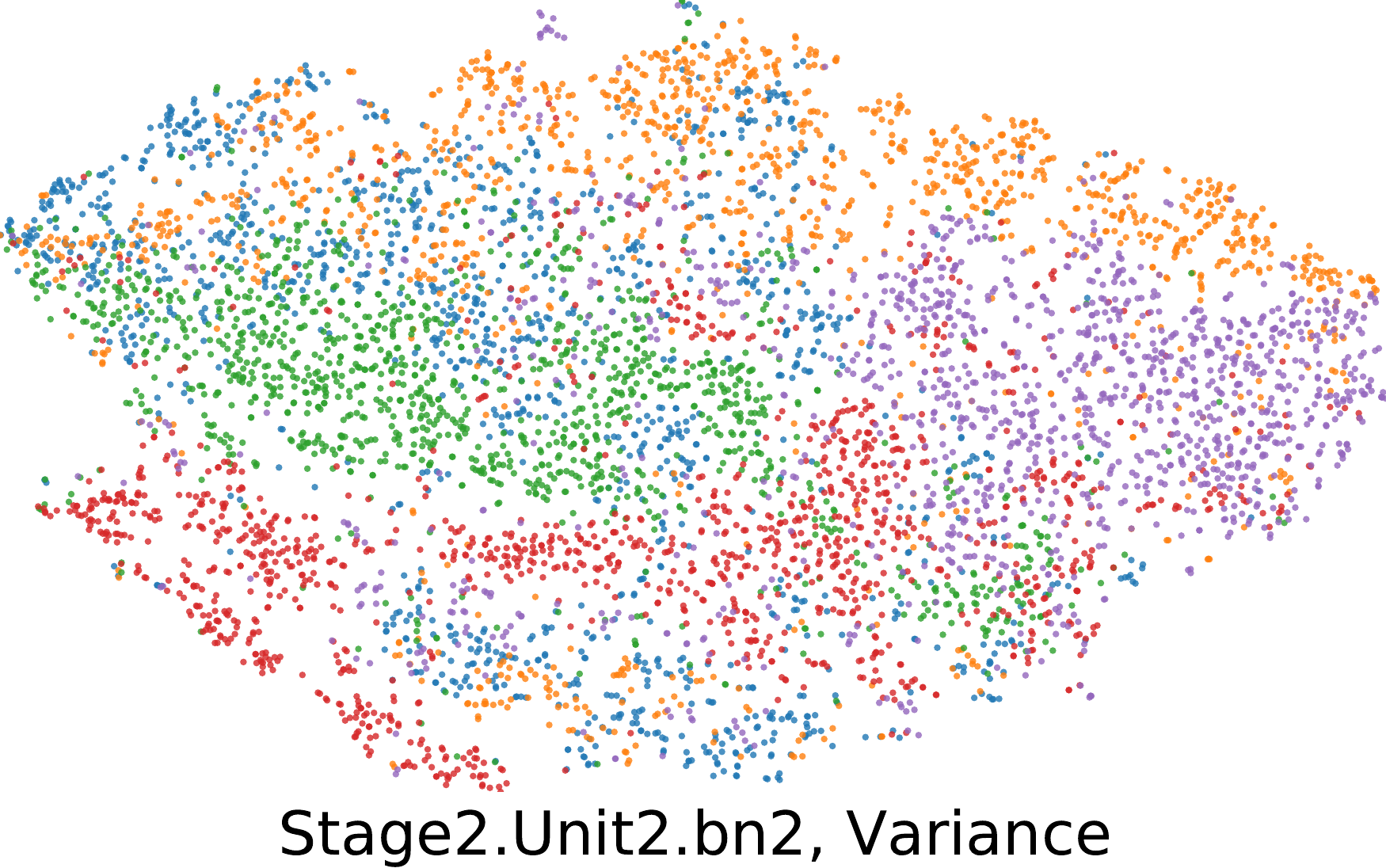} \label{visualization:6} 
}
\subfigure[]{
\includegraphics[height=0.145\linewidth]{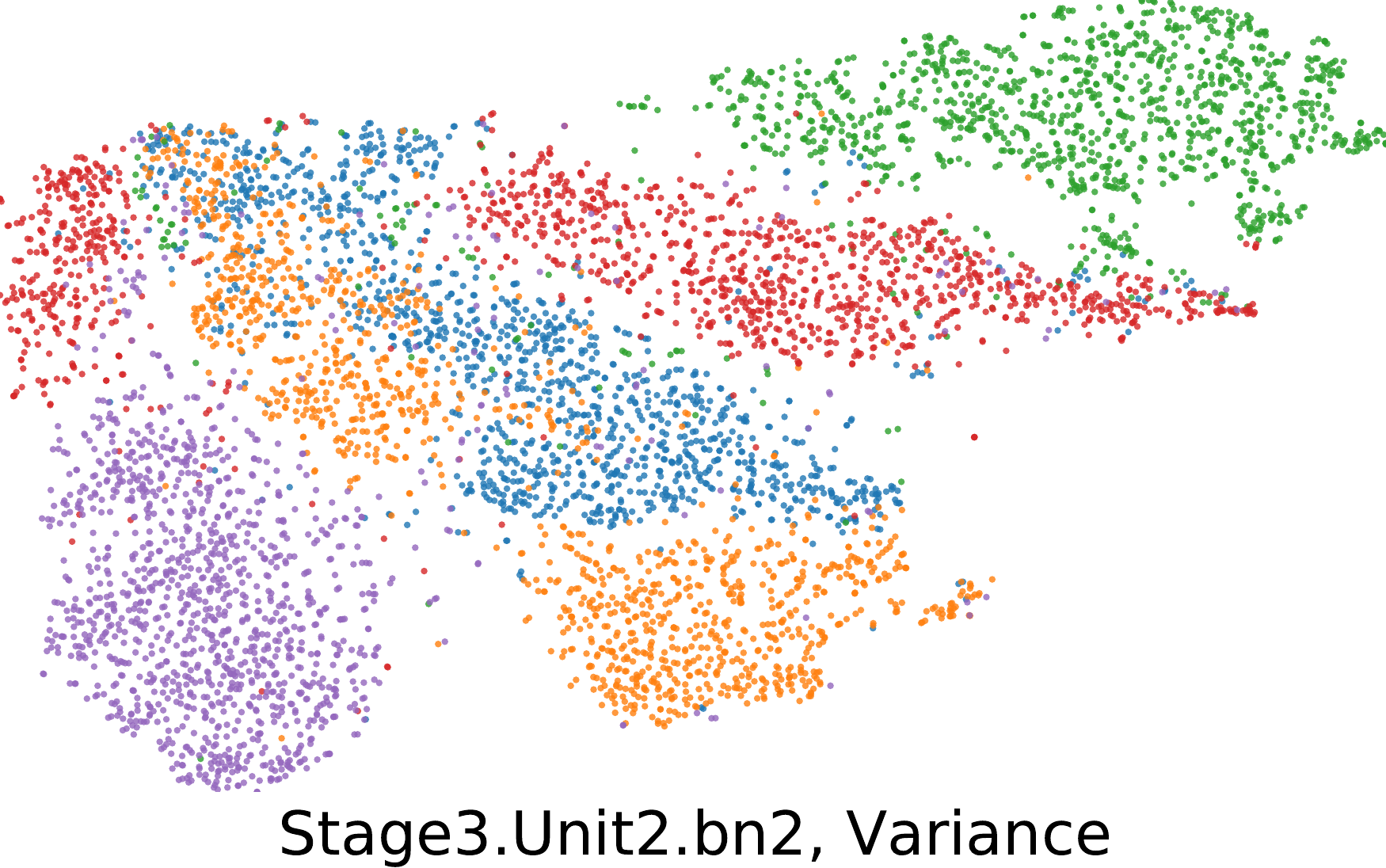} \label{visualization:7} 
}
\subfigure[]{
\includegraphics[height=0.145\linewidth]{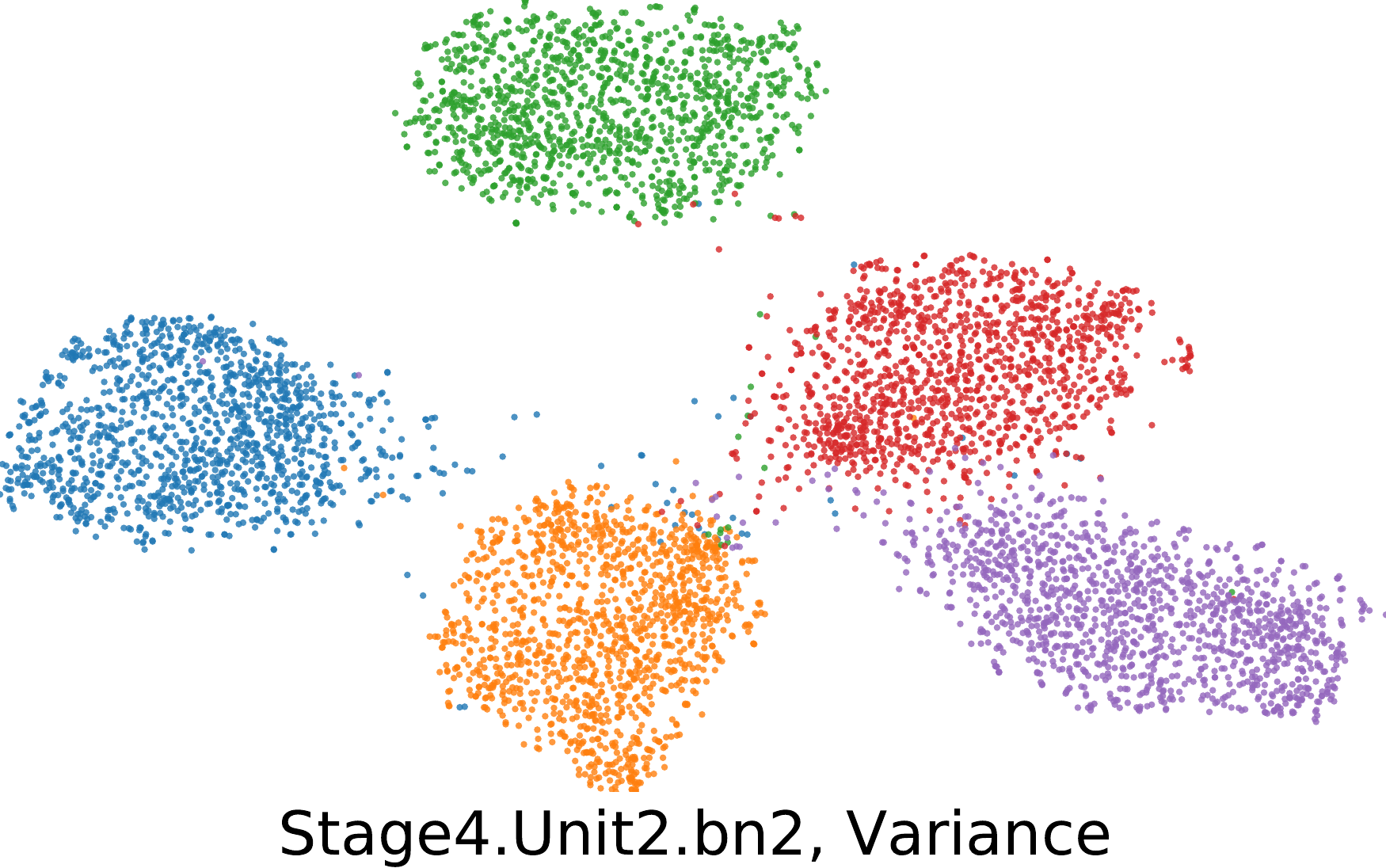}\label{visualization:8}
}
\caption{t-SNE visualization (five classes) of BNS in different layers of pre-trained ResNet-18 on ImageNet. BNS in shallow layers are overlapping. For deep layers, different classes have varying BNS and there exists small distortion within data from the same class. Similar observations can be found in other networks as well (See the supplementary materials). Best viewed in color.}
\label{visualization}
\vspace{-1.5em}
\end{figure*}

Motivated by the above analyses, in this paper, we propose fine-grained data distribution alignment (termed FDDA) for post-training quantization, as illustrated in Fig.\,\ref{framework}. FDDA is inspired by the fact that the representation of synthetic data has been demonstrated to be feasible in many other tasks such as image super-resolution~\cite{zhang2021data}, light-weight student network~\cite{chen2019data} and zero-shot quantization~\cite{GDFQ}. Thus, for the first time, we apply it to post-training quantization to tackle the insufficient data problem, providing a new perspective for post-training quantization. 
Following GDFQ~\cite{GDFQ}, we take full use of the pre-trained model to guide the generator to synthesize fake data. Except for the distillation of output logits between the pre-trained model and its quantized version to improve the quantization performance, GDFQ also retains the distribution information of training data, modeled by the batch normalization statistics (BNS) in the pre-trained model. To this end, the mean and variance of the synthetic data distribution are constrained to be the same as those of the real data distribution.
However, we realize that, based on our two insightful observations on BNS, this information retaining manner is very coarse for deep layers.

Specifically, we calculate the mean and variance $\emph{w.r.t.}$ each image sample and the statistical results are visualized in Fig.\,\ref{visualization}. As can be seen, the BNS of different classes are overlapping in shallow layers. However, two properties of deep BNS including inter-class separation and intra-class incohesion are observed. The former indicates different classes possess varying BNS while the latter indicates a small distortion of BNS among data from the same class. The BNS captured in the pre-trained model only reflects the distribution of the whole dataset, which are applicable to shallow layers with mixed class-wise BNS. However, these BNS in the pre-trained model are very coarse for deep layers with separable class-wise BNS. Thus, the synthetic data needs a fine-grained BNS alignment for these deep layers.

To this end, we further introduce a BNS-centralized loss and a BNS-distorted loss, respectively to align the fine-grained BNS properties of inter-class separation and intra-class incohesion. In contrast to zero-shot quantization~\cite{zeroq}, an additional calibration dataset, usually comprising one image per class, can be available in post-training quantization. To fully utilize this bonus, we derive means and variances of each image in deep layers of the pre-trained model, and then define the computed means and variances as the BN centers of each class. To preserve inter-class separation, our BNS-centralized loss forces the synthetic data distributions of different classes closely to their own centers. To preserve intra-class incohesion, we add Gaussian noise into the centers to mimic the distortion and the BNS-distorted loss forces the synthetic data distribution of the same class closely to the distorted centers. Through a fine-grained BNS alignment, our proposed FDDA significantly improves the quantization performance over existing methods on ImageNet~\cite{li2021brecq,BitSplitStitching,adaquant}, particularly when the first and last layers are also quantized to low-bit. Our contributions are three-fold:

\begin{itemize}
\item To our best knowledge, we are the first to explore combining calibration dataset with synthetic data in post-training quantization, which might provide a new perspective for post-training quantization.
\item We observe properties of inter-class separation and intra-class incohesion in deep BNS. Besides, we devise a BNS-centralized loss and a BNS-distorted loss to preserve these two properties in synthetic data.
\item Extensive experiments demonstrate that our FDDA can well improve the performance on ImageNet. For example, our FDDA outperforms the current SOTA, BRECQ~\cite{li2021brecq}, by $6.64\%$ in the top-1 accuracy when all layers of MobileNet-V1 are quantized to 4-bit.
\end{itemize}

\section{Related Work}
In this section, we briefly discuss the most related work to ours including post-training quantization and zero-shot quantization. A more comprehensive survey is referred to~\cite{gholami2021survey}.

\textbf{Post-training Quantization}. 
Most existing post-training quantization methods attempt to alleviate the accuracy deterioration problem from two perspectives: designing more sophisticated quantization methods and introducing a new rounding function or loss function.
From the first perspective, Liu \emph{et al}.~\cite{MPwMP} proposed to close the gap between the full-precision weight vector and its low-bit version by using the linear combination of multiple low-bit vectors.
Wang \emph{et al}.~\cite{BitSplitStitching} finished quantization in a two-stage manner of bit-split and bit-stitching. In the bit-split stage, the $K$-bit constraint of integer is split into $(K-1)$ ternary learning problems, and each bit is then separately solved in an iterative optimization procedure. In the bit-stitching stage, the $K$-bit integer is recovered by the linear combination with a base of $2^{k-1}$ for the $k$-th bit. In addition, they also use channel-wise quantizer for activations and integrate the scaling factor into corresponding 2D kernels to avoid extra storage.
Piece-wise linear quantization~\cite{fang2020post} splits the whole weights into two non-overlapping areas, including one dense region comprising low-magnitude weights, and one sparse region comprising high-magnitude weights. On top of the splitting, both areas are respectively quantized into the same low-bit. 
To model the quantization parameters, a linear regressor is constructed to predict the $\alpha$-quantile of activations~\cite{kryzhanovskiy2021qpp}, which eliminates the involvement of complex sorting algorithm.
From the second perspective, AdaRound~\cite{Upordown} analyzes that it is not advisable to simply round full precision weight to its nearest fixed-point value. Alternatively, the rounding problem is formulated as a per-layer quadratic unconstrained binary optimization problem, based on which, a continuous relaxation is introduced to find an adaptive rounding.  BRECQ~\cite{li2021brecq}, one of the state-of-the-art methods, builds a block-wise reconstruction between the outputs of the full precision network and quantized network to achieve a balance between cross-layer dependency and generalization error. Besides, trainable clipping~\cite{PACT} for activations is also considered by BRECQ. Similar motivation can also be found in earlier works~\cite{APTQ,BitSplitStitching,ACIQ}.

\textbf{Zero-Shot Quantization}.
Zero-shot quantization is not permitted to access the training dataset. Thus, synthetic samples become an alternative to calibrate and fine-tune the quantized models. According to the methodology of data synthesis, we categorize existing studies into two groups: data optimizer~\cite{yin2020dreaming,DSG,zeroq,Theknowledgewithin} and data generator~\cite{GDFQ,ZAQ,choi2021qimera}. 
Data optimizer based methods produce synthetic images from Gaussian noise~\cite{yin2020dreaming}. 
ZeroQ~\cite{zeroq} forwards propagated the Gaussian inputs to collect BNS. Then the optimization that minimizes the difference between the collected BNS and the BNS in the pre-trained model is constructed to update the Gaussian inputs to ensure that the synthetic data does not deviate from the real data distribution.
Except for aligning the BNS, a Domain Prior loss and an Inception loss are introduced in~\cite{Theknowledgewithin}. The former encourages nearby pixels between the input image and its Gaussian-smoothed variant to be similar. The latter prevents the model from producing inputs that lead to exploding outputs.
To break the data homogenization, DSG~\cite{DSG} slacks the alignment of BNS and introduces a layer-wise enhancement to enhance diverse data samples.
As for data generator, this group is featured with a generator in Generative Adversarial Networks (GAN)~\cite{GAN} to synthesize images.
GDFQ~\cite{GDFQ} exploits the classification boundary knowledge and distribution information in the pre-trained model, and then devises a knowledge matching generator to produce synthetic data for model quantization.
To diversify generated data, ZAQ~\cite{ZAQ} trains the quantized model and generator in an adversarial fashion by adopting an elaborated two-level discrepancy. 
To capture the distribution of the original data lies on the decision boundaries, Qimera~\cite{choi2021qimera} introduces superposed latent embeddings to produce boundary supporting samples.

\section{Methodology}\label{Method}

\subsection{Preliminaries}

\subsubsection{Quantizer.} Following~\cite{zeroq,GDFQ}, we adopt asymmetric uniform quantization in this paper. Given the data $\bm{x}$ (weights or activations), bit-width $b$, lower bound $l$ and upper bound $u$, the quantizer is defined as:
\begin{equation}
\mathbf{q} = round(\frac{clip(\bm{x}, l, u)}{s}),
\label{Equation1}
\end{equation}
where $clip(\bm{x}, l, u) = min\big(max(\bm{x}, l), u\big)$, $round(\cdot)$ rounds its input to the nearest integer, $s = \frac{u - l}{2^b-1}$ is the scaling factor that projects a floating-point number to a fixed-point integer, and $\mathbf{q}$ is the quantized fixed-point number. The corresponding de-quantized item $\bar{\bm{x}}$ can be obtained as:
\begin{equation}
\bar{\bm{x}} = \mathbf{q} \cdot s.
\end{equation}

We use layer-wise quantizer and channel-wise quantizer for activations and weights, respectively. The lower bound $l$ and upper bound $u$ are set to the minimum and maximum of per-layer activations (per-channel weights).

\subsubsection{Data Synthesis.} 
Ideally, post-training quantization completes network compression with a small calibration dataset $D = \{ (\hat{\bm{\bm{x}}}, \hat{y}) \}$, typically consisting of one image per class\footnote{Occasionally, the label $\hat{y}$ is not available. In this case, it can be predicted by the pre-trained full-precision model.}. However, this small calibration dataset fails to retain performance when quantizing the network to very low precision, such as 4-bit. Inspired by zero-shot quantization, we resort to data synthesis.
As shown in Fig.\,\ref{framework}, we deploy a generator $G$ to synthesize an image $\tilde{\bm{x}}$ from a random Gaussian noise $\tilde{\mathbf{z}}$ conditioned on the target label $\tilde{y}$, \emph{i.e.}, $\tilde{\bm{x}} = G(\tilde{\mathbf{z}}|\tilde{y})$. We expect that the synthetic data to be similar to the real data. Despite the inaccessibility of the whole training data, we can turn to the data distribution information captured by the batch normalization statistics (BNS) in the pre-trained model $F$. Following~\cite{GDFQ}, the BNS loss can be adopted to preserve the distribution:
\begin{equation}
{\cal L}^G_{\text{BNS}} = \sum_{l=1}^{L} \|\bm{\mu}'_l(\tilde{\bm{\bm{x}}})-\bm{\mu}_l^F \|^2 + \| \bm{\sigma}'_l(\tilde{\bm{\bm{x}}})-\bm{\sigma}_l^F \|_2^2,
\label{bns}
\end{equation}
where $\bm{\mu}_l^F$ and $\bm{\sigma}_l^F$ are the running mean and variance in the $l$-th layer of pre-trained $F$. $\bm{\mu}'_l(\cdot)$ and $\bm{\sigma}'_l(\cdot)$ return the mean and variance of input data in the $l$-th layer of $F$.

\subsubsection{Classification.}
%
We also use cross-entropy loss to ensure synthetic data can be correctly classified by the pre-trained model $F$:
\begin{equation}
{\cal L}^G_{\text{CE}} = \mathbb{E}_{(\bm{x}, y) \sim \{(\tilde{\bm{x}}, \tilde{y})\}}\big[\text{CE}\big(F(\bm{x}), y\big)\big].
\label{ce_g}
\end{equation}

Note that, we fix $F$ during the whole training process, and the generator $G$ is updated instead.

 %
%

\begin{figure}[t]
\centering
\includegraphics[height=0.3\linewidth]{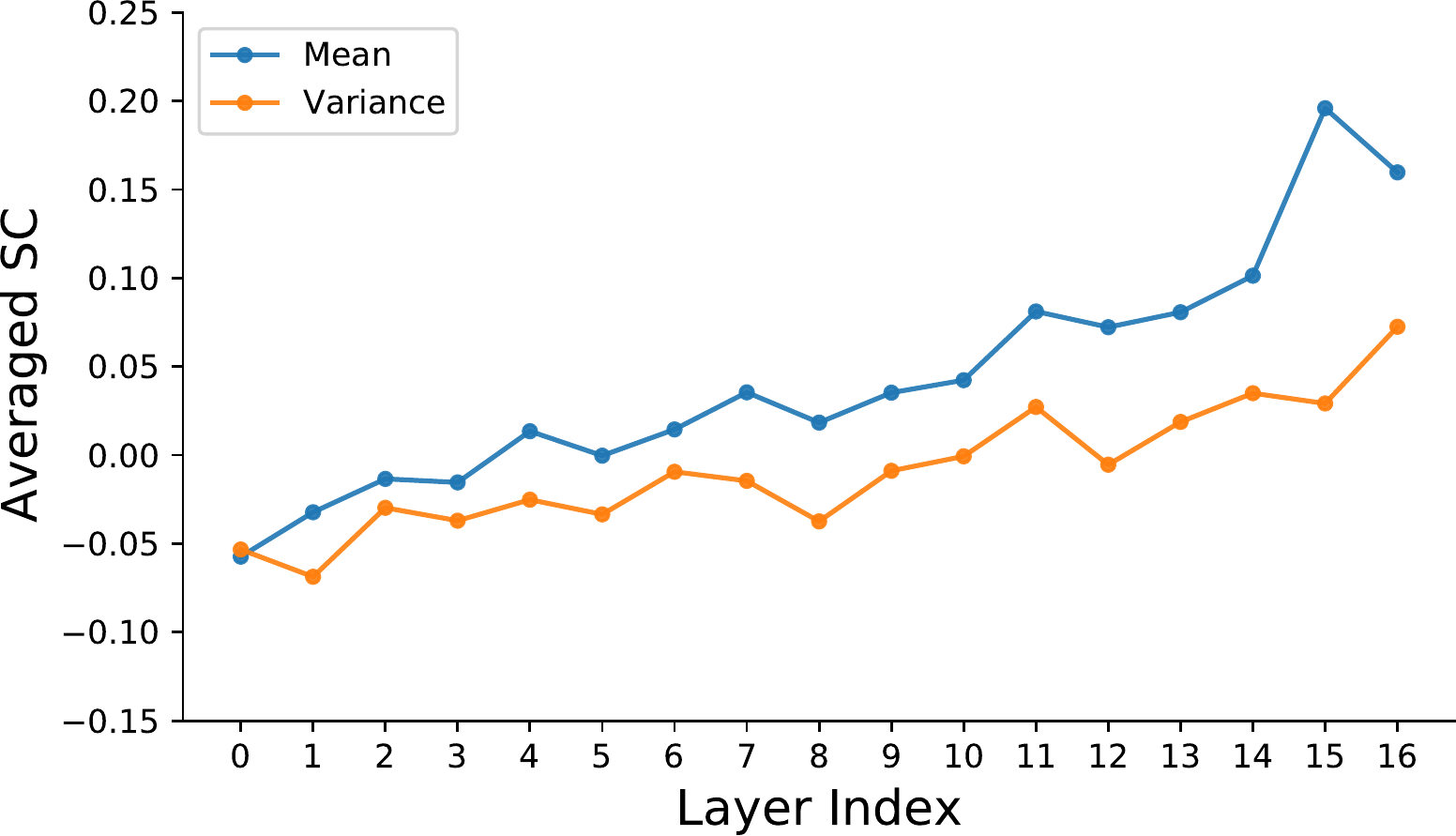}
\caption{The Average of Silhouette Coefficient values \emph{w.r.t.} the BNS in different layers.}
\label{sc}
\end{figure}

\subsection{Our Insights}
The BNS in pre-trained $F$ are calculated by a series of averages of different batches of the full training set, \emph{a.k.a.}, moving average. As a consequence, they capture the distribution of the whole dataset. However, it is unclear whether they can be a representative of per-class BNS or even per-image BNS. To verify this, we feed the whole ImageNet to ResNet-18~\cite{he2016deep} and calculate per-image mean vector $\bm{\mu}'_l$ and variance vector $\bm{\sigma}'_l$ in each layer. Fig.\,\ref{visualization} visualizes several examples via t-SNE~\cite{tSNE}.

As can be seen, the BNS over different classes vary a lot across different layers. Specifically, the BNS of different classes are overlapping in shallow layers while there is regularity in deep layers, which, we outline into two properties: \textbf{inter-class separation} and \textbf{intra-class incohesion}. The former indicates that the BNS within the same class are formed into one cluster and different classes are separable. The latter refers to a small distortion of BNS among data from the same class. To quantitatively measure these two properties, we introduce Silhouette Coefficient (SC)~\cite{rousseeuw1987silhouettes}, a value of which reflects how similar an object is to its own cluster (cohesion) in comparison with other clusters (separation). SC value of one sample $\mathbf{v}$ is defined as:
\begin{equation}
\text{SC}(\mathbf{v}) = \frac{{b(\mathbf{v})-a(\mathbf{v})}}{max\big(a(\mathbf{v}), b(\mathbf{v})\big)},
\label{Silhouette_Coefficient_equation}
\end{equation}
where $a(\mathbf{v})$ denotes the average of intra-cluster distance for sample $\mathbf{v}$, and $b(\mathbf{v})$ is the average of nearest-cluster distance for sample $\mathbf{v}$. Note that $b(\mathbf{v})$ is the distance between $\mathbf{v}$ and the nearest cluster that $\mathbf{v}$ is not a part of. The value of $\text{SC}(\mathbf{v})$ ranging from -1 (separation) to +1 (cohesion). Values near 0 indicate overlapping clusters. Negative values generally indicate that a sample has been assigned to the wrong cluster, as a different cluster is more similar.
With Eq.\,(\ref{Silhouette_Coefficient_equation}), we can obtain SC values for each mean vector $\bm{\mu}'_l$ and variance vector $\bm{\sigma}'_l$. Fig.\,\ref{sc} displays the average of all input samples in each layer. As can be seen, the SC values are very small, even negative, in shallow layers, which indicates overlapping clusters. On the contrary, SC values increase in deep layers, which indicates inter-class separation. However, the increase of SC is very limited (no more than 0.25), which indicates a relatively large intra-cluster distance, \emph{i.e.}, intra-class incohesion. These analyses are consistent with the observations in Fig.\,\ref{visualization}.

\subsubsection{In-depth Analysis.}
The observed inter-class separation and intra-class incohesion can be explained by the fact that networks extract class-unrelated universal low-level features in shallow layers such as edges and curves. While in deep layers, networks are learned to extract class-related semantic features distinguishable from other classes in deep layers, leading to inter-class separation. The intra-class incohesion results from the varying image contents though these images are from the same class. For shallow layers, the overlapping clusters hardly model a subtle per-class distribution, thus the BNS in the pre-trained model can be an alternative. However, the BNS in the pre-trained model are very coarse and the constraint of Eq.\,(\ref{bns}) cannot model the properties of per-class separation and intra-class incohesion. Thus, in addition to the coarse-grained alignment, a fine-grained BNS alignment is also necessary for deep layers.

\subsection{Fine-grained BNS Alignment}

To preserve the properties of inter-class separation and intra-class incohesion, in this subsection, we respectively introduce a BNS-centralized loss and a BNS-distorted loss. Details are presented below.


\subsubsection{BNS-centralized Loss.} 
Since the BNS of each class are formed into one cluster, we can place a centroid as an explicit supervisory signal for each class and force the per-class distribution of synthetic images to be close to the assigned centroid. Recall that a real image $(\hat{\bm{\bm{x}}},\hat{y})$ per class can be available from the calibration dataset $D$ in post-training quantization. It is reasonable to use the BNS of image $\hat{\bm{\bm{x}}}$ as the corresponding centroid of class $\hat{y}$ since its BNS already fall into the target cluster and are separable from BNS of other classes. To this end, given $\hat{\bm{\bm{x}}}$ with its label $\hat{y} = c$, we define the following BNS-centralized loss such that synthetic images can be further aligned to their corresponding centroids:
\begin{equation}
\begin{split}
{\cal L}^G_{\text{C-BNS}} = &\sum_{l=K}^{L}  \|\bm{\mu}'_l(\tilde{\bm{x}}|\tilde{y} = c)-\bm{\mu}'_l(\hat{\bm{x}}|\hat{y} = c) \|^2 \\&
+ \| \bm{\sigma}'_l(\tilde{\bm{x}}|\tilde{y} = c)-\bm{\sigma}'_l(\hat{\bm{x}}|\hat{y} = c) \|^2,
\label{c-bns}
\end{split}
\end{equation}
where $K$ is a pre-given hyper-parameter denoting the start of deep layers. In all experiments, we set $K = \text{ceil}(\frac{L}{2}) - 2$ where $\text{ceil}(\cdot)$ is the rounding up function.

\subsubsection{BNS-distorted Loss.} Our BNS-centralized loss ensures the fine-grained separableness across different classes. However, how to retain the incohesion within the same class remains an issue. To solve this, we further propose to distort the centroid of per-class BNS by introducing Gaussian noise and define the following BNS-distroted loss:
\begin{equation}
\begin{split}
{\cal L}^G_{\text{D-BNS}}=&\sum_{l=K}^{L} \big\| \bm{\mu}'_l(\tilde{\bm{x}}|\tilde{y} = c) -\mathcal{N}\big(\bm{\mu}'_l(\hat{\bm{x}}|\hat{y} = c), \bm{\upsilon_{\mu}}\big) \big\|^2 \\& + 
\big\| \bm{\sigma}'_l(\tilde{\bm{x}}|\tilde{y} = c)-\mathcal{N}\big(\bm{\sigma}'_l(\hat{\bm{x}}|\hat{y} = c), \bm{\upsilon_{\sigma}}\big) \big\|^2,
\end{split}
\label{d-bns}
\end{equation}
where $\bm{\upsilon_{\mu}} = 0.5$ and $\bm{\upsilon_{\sigma}} = 1.0$ are used to control the distortion degrees of mean and variance. For each synthetic data $\tilde{\bm{x}}$, its target is sampled from a Gaussian distribution centered on the class centroid of $\tilde{\bm{x}}$. As a result, our BNS-distorted loss provides diverse distorted centroids which prevent the BNS of per synthetic data from overfitting its centroid. And by doing this, we can further retain the intra-class incohesion.

Our experimental results in Sec.\,\ref{ablation} show that the BNS-distorted loss can retain the inter-class separation to some extent since the distorted centroid for synthetic data is centered on the corresponding class centroid. However, it is hard to manually model the Gaussian noise exactly such that the inter-class separableness and intra-class incohesion can be well preserved in the synthetic images simultaneously. Thus, both BNS-centralized loss and BNS-distorted loss are necessary as verified in the experiment.



%

\subsection{Model Quantization}
\subsubsection{Classification.}
To take full use of the available data, both calibration images and synthetic images are used to fine-tune the quantized model $Q$, which can be realized through the cross-entropy loss:
\begin{equation}
{\cal L}^Q_{\text{CE}} = \mathbb{E}_{(\bm{x}, y) \sim D \cup \{(\tilde{\bm{x}}, \tilde{y})\}}\big[\text{CE}\big(Q(\bm{x}), y\big)\big].
\label{ce_q}
\end{equation}

\subsubsection{Distillation.} 
It is possible that the synthetic image does not include corresponding class-specific features. As a result, $\tilde{y}$ may be unreliable. Thus, we apply knowledge distillation (KD)~\cite{hinton2015distilling} to transfer the outputs of full-precision model $F$ to quantized model $Q$ so that even though the synthetic image may have an inaccurate label, $Q$ can still be correctly optimized by learning the soft target provided by $F$. Moreover, KD is also beneficial to the learning of calibration data. The KD loss is defined by the Kullback-Leibler distance $\text{KL}(\cdot,\cdot)$ as:
%
\begin{equation}
{\cal L}_{\text{KD}}^Q = \mathbb{E}_{(\bm{x}, y) \sim D\cup \{(\tilde{\bm{x}}, \tilde{y})\}}\big[\text{KL}\big(Q(\bm{x}), F(\bm{x})\big)\big].
\label{kd_q}
\end{equation}

\subsection{Training Process}
The training of our method consists of updating the generator $G$ and the quantized model $Q$, where $Q$ is obtained by quantizing the pre-trained full-precision model $F$. $G$ produces a set of synthetic images while $Q$ is trained with the aid of synthetic images and calibration images. We also emphasize that the $F$ is fixed without any updating during the whole training process.

\subsubsection{Updating Generator $G$.} 
With a random Gaussian noise $\tilde{\bm{z}}$ conditional on label $\tilde{y}$ as its input, the generator $G$ synthesize an image $\tilde{\bm{x}} = G(\tilde{\bm{z}}|\tilde{y})$, which is then used for classification of Eq.\,(\ref{ce_g}) and preserving distribution of training set including the coarse alignment of Eq.\,(\ref{bns}), inter-class separation of Eq.\,(\ref{c-bns}) and intra-class incohesion of Eq.\,(\ref{d-bns}). Thus, the overall loss for the generator $G$ is derived as:
\begin{equation}
\begin{split}
\begin{aligned}
{\cal L}^G =& \alpha_1 \cdot {\cal L}^G_{\text{CE}}+\alpha_2 \cdot {\cal L}^G_{\text{BNS}} +\alpha_3 \cdot {\cal L}^G_{\text{D-BNS}} + \alpha_4 \cdot \mathcal{L}^G_{\text{C-BNS}},
\end{aligned}
\end{split}
\label{generator}
\end{equation}
where the $\alpha_1, \alpha_2, \alpha_3$ and $\alpha_4$ are the trade-off parameters.



\subsubsection{Updating Quantized Model $Q$.} 
The quantized model $Q$ takes the synthetic data and calibration data as its inputs, and then the classification loss of Eq.\,(\ref{ce_q}) and distillation loss of Eq.\,(\ref{kd_q}) are constructed to retain the performance. Thus, the overall loss for the quantized model $Q$ is derived as:
\begin{equation}
{\cal L}^Q = {\cal L}_{\text{CE}}^Q+\alpha_5 \cdot {\cal L}_{\text{KD}}^Q,
\label{quantization}
\end{equation}
where $\alpha_5$ is a trade-off parameter.


\section{Experimentation}\label{experimentation}

\subsection{Implementation Details}
We choose to quantize ResNet-18~\cite{he2016deep}, MobileNetV1~\cite{howard2017mobilenets}, MobileNetV2~\cite{sandler2018mobilenetv2} and RegNet-600MF~\cite{regnetx}. All experiments are conducted on the challenging ImageNet with 1.2 million training images and 50,000 validation images from 1,000 classes~\cite{russakovsky2015imagenet}. The calibration dataset consists of 1,000 images including one image per class. We report the top-1 accuracy and the code is implemented using  Pytorch~\cite{paszke2019pytorch}.

For ease of implementation, we directly import the generator from GDFQ~\cite{GDFQ} to produce synthetic images.
The initial learning rates for the generator and quantized network are set to $10^{-3}$ and $10^{-6}$ respectively. 
For the generator, the optimizer is Adam~\cite{kingma2014adam} with $0.9$ as the momentum and the learning rate are multiplied by 0.1 every 100 epochs. 
For the quantized network, the optimizer is SGD with Nesterov~\cite{nesterov1983method} with $10^{-4}$ as the weight decay and we adjust the learning rate using the cosine annealing~\cite{loshchilov2016sgdr}.
%
%
Before formal training, we set up a warm-up updating of the generator $G$ for 50 epochs. Then, a total of 350 epochs are used to update the generator $G$ and quantized model $Q$.
%

\subsection{Experimental Results}

\begin{minipage}{\textwidth}
    \centering
    \begin{minipage}[t]{0.45\textwidth}
    \centering
        \makeatletter\def\@captype{table}\makeatother\caption{Comparison with GDFQ~\cite{GDFQ} by quantizing all layers of ResNet-18 to 4-bit. TAQ denotes training-aware quantization. ``C'' indicates the calibration dataset. ``F'' represents fine-grained data distribution alignment. Note that GDFQ + C + F = FDDA.} \vspace{1.2em}
        \begin{tabular}{c|c}
            \hline
            Method            & Acc. (\%) \\ \hline
            Full precision    & 71.47     \\ \hline
            TAQ      & 68.24     \\ \hline
            GDFQ   & 60.60\\
            GDFQ + C            & 65.64     \\
            GDFQ + C + F (\textbf{FDDA})   & \textbf{68.88}       \\
            \textbf{FDDA} + w/o label    & \textbf{68.68}   \\ \hline
        \end{tabular}
    \label{baseline}
    \end{minipage} \hspace{0.5em}
\begin{minipage}[t]{0.45\textwidth}
   \centering
    \makeatletter\def\@captype{table}\makeatother\caption{Comparison with zero-shot methods when all layers of ResNet-18 are quantized to 4-bit. ``C'' indicates the calibration dataset.} \vspace{1.2em}
        \begin{tabular}{c|c}
            \hline
            Method            & Acc. (\%) \\ \hline
            Full precision    & 71.47     \\ \hline
            DI~\cite{yin2020dreaming} + C & 65.68 \\
            ADI~\cite{yin2020dreaming} + C & 66.30 \\
            ZeroQ~\cite{zeroq} + C & 66.89 \\
            ZAQ~\cite{ZAQ} + C & 29.50 \\
            GDFQ~\cite{GDFQ} + C      & 65.64     \\
            Qimera~\cite{choi2021qimera} + C & 66.19 \\
            DSG~\cite{DSG} + C & 66.83 \\
            \textbf{FDDA} (Ours)   & \textbf{68.88}       \\ \hline
        \end{tabular}
    \label{zero-shot}
    \end{minipage}
\end{minipage}

\subsubsection{Comparison with Zero-Shot Methods.}

We first compare with the zero-shot GDFQ~\cite{GDFQ}, since the data synthesis of our FDDA is built upon the framework of GDFQ. 
Table\,\ref{baseline} displays our experimental results when all layers of ResNet-18~\cite{he2016deep} are quantized to 4-bit. The accuracy of the full-precision model decreases from 71.47\% to around 68.24\% when the training-aware quantization, which requires all training data, is applied. 
Regarding GDFQ which only considers the synthetic data for fine-tuning the quantized model, performance severely degenerates to 60.60\%. Such poor performance disables the application of GDFQ. Given the calibration dataset, GDFQ can increase to 65.48\%, well demonstrating the correctness of our motive in combining real calibration dataset with synthetic data. Nevertheless, the coarse BNS alignment in GDFQ fails to model the fine-grained properties of inter-class separableness and intra-class incohesion. In contrast, our FDDA increases the performance to 68.88\%, better than the training-aware quantization. This result shows the efficacy of our BNS-centralized loss and BNS-distorted loss in synthesizing better images.

In our settings, we assume to have access to the image labels. However, these labels are sometimes not available in real applications. Luckily, the pre-trained full-precision model can be used to predict these labels. In Table\,\ref{baseline}, we also report the performance of our FDDA, \emph{i.e.}, 68.68\%, using predicted labels. The slight drops are attributed to some of the misclassified labels. Nevertheless, our FDDA without real labels still maintains better performance than GDFQ with real images, well demonstrating the importance of preserving the fine-grained inter-class separableness and intra-class incohesion in learning to synthesize images.

Table\,\ref{zero-shot} further shows the comparison between our FDDA with advanced zero-shot studies including data optimizer based methods~\cite{zeroq,DSG,yin2020dreaming} and data generator based methods~\cite{GDFQ,ZAQ,choi2021qimera}. For the former group, we use the calibration dataset as well as 10,000 synthetic images to train the quantized model, and the training process is the same with ours. For the latter group, we insert the calibration dataset into their training process. In Table\,\ref{zero-shot}, when the calibration dataset is applied to all methods, our FDDA still outperforms the advance of DSG by a large margin of 2.05\%, which again demonstrates the efficacy of our BNS-centralized loss and BNS-distorted loss.

\begin{table*}[!t]
\centering
\caption{Comparisons with existing post-training quantization methods. WBAB indicates the weights and activations are quantized to B-bit while FBLB indicates the first layers and last layers are quantized to B-bit. 
%
%
}
\begin{tabular}{c|c|c|c|c|c}
\hline
\multicolumn{1}{l|}{} & \cellcolor[HTML]{FFFFFF}{\color[HTML]{333333} Methods} & ResNet-18 & MobileNetV1 & MobileNetV2 & RegNet-600MF \\ \hline
\multicolumn{1}{c|}{Settings}         & Full precision   & 71.47     &    73.39         &   72.49          &          73.71             \\ \hline
                              & ACIQ-Mix~\cite{ACIQ} &  68.34 &     52.34        &     61.74        &      69.53      \\
                              & AdaQuant~\cite{adaquant}   &    68.56       &      -   &    65.19         &       -   \\
                              & Bit-Split~\cite{BitSplitStitching}   &    69.10       &   -   &  - & -                \\
                              & BRECQ~\cite{li2021brecq}  &    70.60       &     70.16   &     70.83    & 73.38       \\
\multirow{-5}{*}{W5A5, F8L8}  & \textbf{FDDA}(Ours)  &   \textbf{70.86}  & \textbf{71.16}& \textbf{71.99}  &  \textbf{73.99} \\ \hline
                              & ACIQ-Mix~\cite{ACIQ} &    67.0     &    5.06      &        39.49&  54.22   \\
                              & AdaQuant~\cite{adaquant}        &     67.50      &  -     &    34.95   &         -     \\
                              & Bit-Split~\cite{BitSplitStitching}   &    67.56       &    -     &      -   &        -    \\
                              & BRECQ~\cite{li2021brecq}     &    69.60   &    63.66    &    66.57   & 68.33          \\
\multirow{-5}{*}{W4A4, F8L8}  & \textbf{FDDA}(Ours)  &  \textbf{69.76} & \textbf{65.76} &\textbf{69.32}  & \textbf{70.33}  \\ \hline
                              & ACIQ-Mix~\cite{ACIQ} &  66.80 &     51.65        &     60.42        &      69.13      \\
                              & AdaQuant~\cite{adaquant}   &    68.19       &      -   &      63.61       &       -   \\
                              & Bit-Split~\cite{BitSplitStitching}   &    68.88       &   -   &  - & -                \\
                              & BRECQ~\cite{li2021brecq}  &    70.27       &     66.51   &     70.26    & 72.78       \\
\multirow{-5}{*}{W5A5, F5L5}  & \textbf{FDDA}(Ours) & \textbf{70.56}  & \textbf{70.26}& \textbf{71.63}  &  \textbf{73.62} \\ \hline
                              & ACIQ-Mix~\cite{ACIQ} &  57.47 &     4.68        &     34.84        &      51.74      \\
                              & AdaQuant~\cite{adaquant}   &    63.45       &      -   &      34.64       &       -  \\
                              & Bit-Split~\cite{BitSplitStitching}   &    67.49       &   -   &  - & -                \\
                              & BRECQ~\cite{li2021brecq}  &    67.94       &     57.11   &     63.64    & 66.17       \\
\multirow{-5}{*}{W4A4, F4L4}  & \textbf{FDDA}(Ours)  &   \textbf{68.88}  & \textbf{63.75}&     \textbf{68.38}   &  \textbf{68.96} \\ \hline
\end{tabular}
\label{comparison}
\end{table*}

\subsubsection{Comparison with Competitors.}
We compare with the recent studies on post-training quantization~\cite{ACIQ,adaquant,BitSplitStitching,li2021brecq}. The quantized networks include ResNet-18~\cite{he2016deep}, MobileNetV1~\cite{howard2017mobilenets}, MobileNetV2~\cite{sandler2018mobilenetv2} and RegNet-600MF~\cite{regnetx}. All networks are quantized to the low precision 5-bit and 4-bit. Besides, to show the advantage of our FDDA, we quantize the first and last layers to 8-bit and lower precision (5-bit or 4-bit).
Table\,\ref{comparison} shows the experimental results.

When the first and last layers of full-precision models are quantized to 8-bit (F8L8), both our FDDA and recent SOTA BRECQ~\cite{li2021brecq} can retain a high performance of the full-precision models regardless of 5-bit (W5A5) or 4-bit (W4A4) weights and activations in other layers. Comparing to BRECQ, our FDDA obtains performance gains by 0.26\%, 1.00\%, 1.16\% and 0.61\% when quantizing ResNet-18, MobileNetV1, MobileNetV2 and RegNet-600MF to W5A5, while they are 0.16\%, 2.10\%, 2.75\% and 2.00\% when quantized to W4A4. We observe that our FDDA retains better performance than BRECQ when quantizing light-weight models such as MobileNets, particularly when lower precision, such as 4-bit, is performed.

When the first and last layers of full-precision models are quantized to lower precision (F5L5 or F4L4) as well, we notice that our FDDA outperforms BRECQ by margins. Specifically, our FDDA increases the performance of BRECQ in W5A5 by 0.29\%, 3.75\%, 1.37\% and 0.84\% \emph{w.r.t.} ResNet-18, MobileNetV2, MobileNetV2 and RegNet-600MF, and the performance gains are 0.94\%, 6.64\%, 4.74\%, 2.79\% in the case of W4A4. These results well verify our statement in the introduction section that the performance improvements of existing studies are usually built on the premise that the first and last layers are quantized to higher precision, and also demonstrates the effectiveness of our combining real calibration dataset with synthetic data to enhance the performance of post-training quantization.

\begin{figure*}[!t]
\centering
\subfigure[]{
\includegraphics[width=0.19\columnwidth]{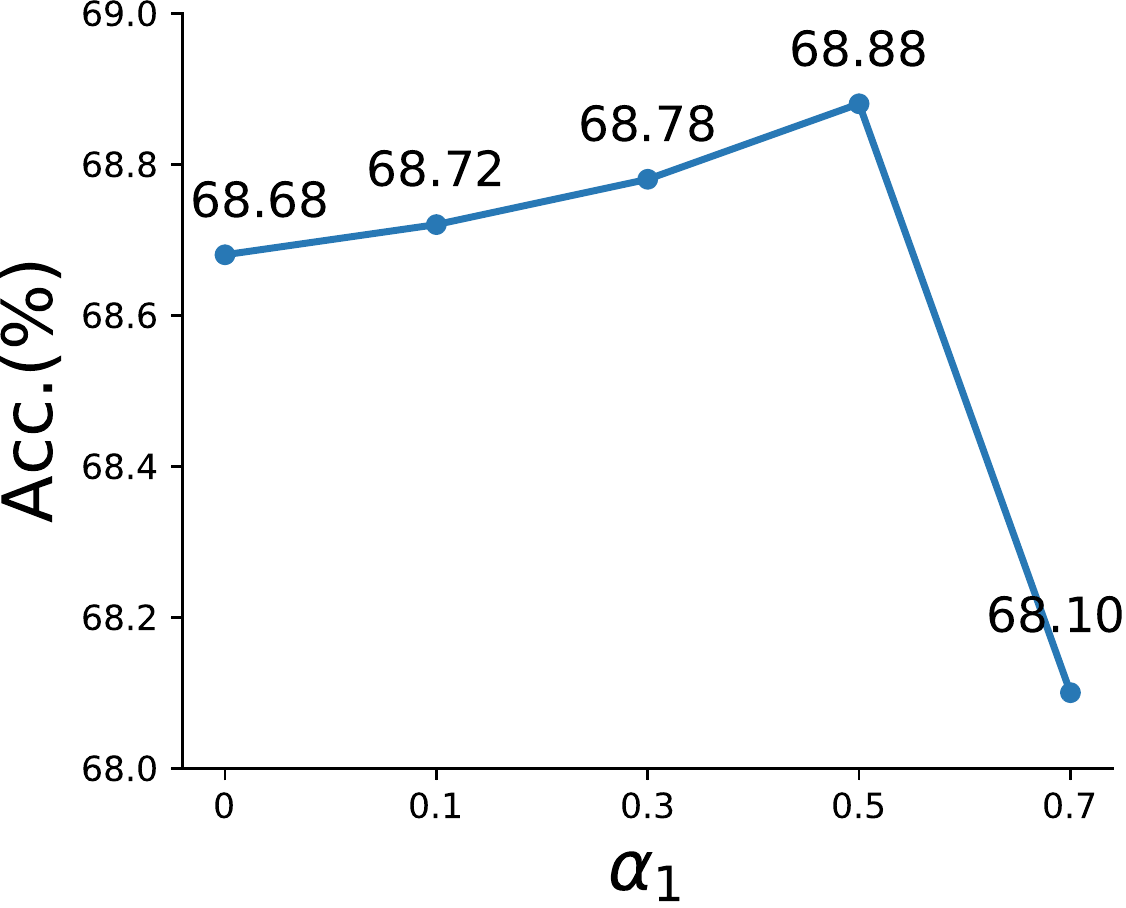} \label{hyperparameters:alpha1}
}\hspace{-3mm}
\subfigure[]{
\includegraphics[width=0.19\columnwidth]{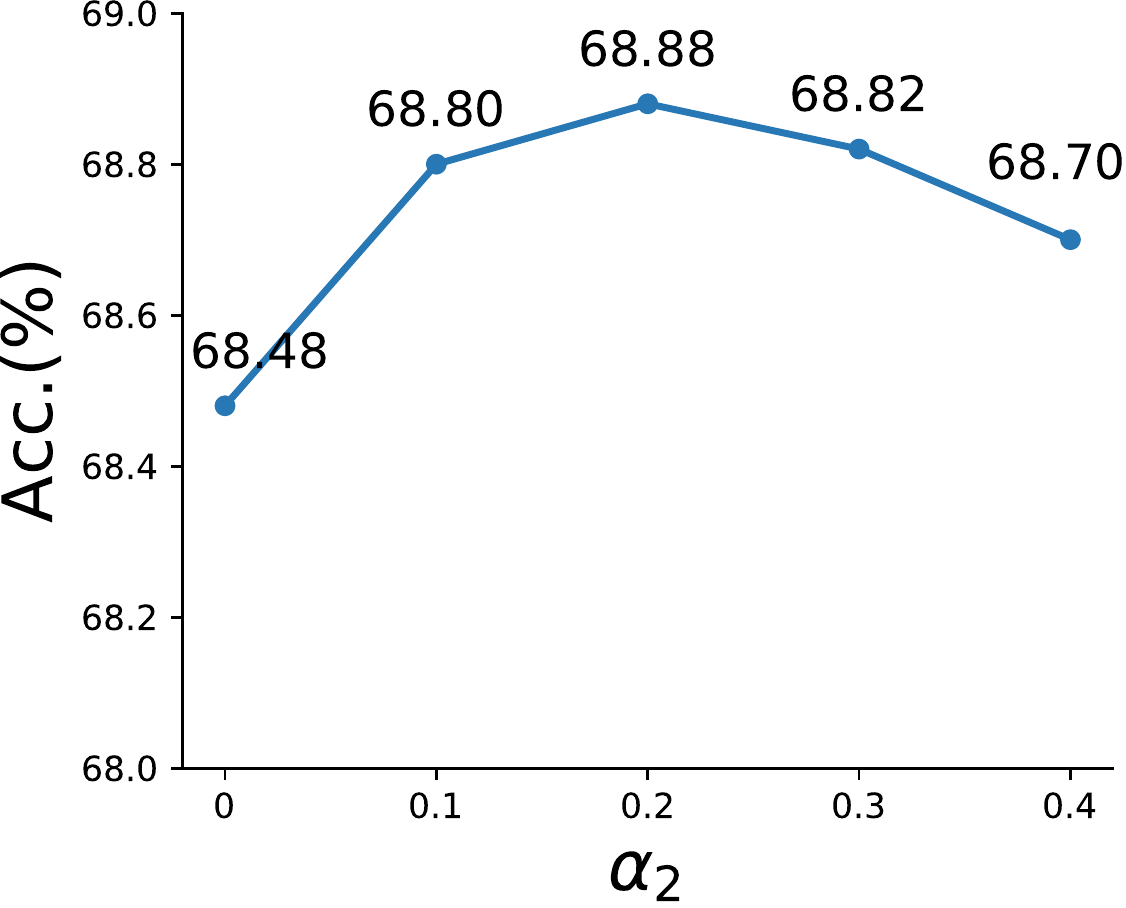} \label{hyperparameters:alpha2}
}\hspace{-3mm}
\subfigure[]{
\includegraphics[width=0.19\columnwidth]{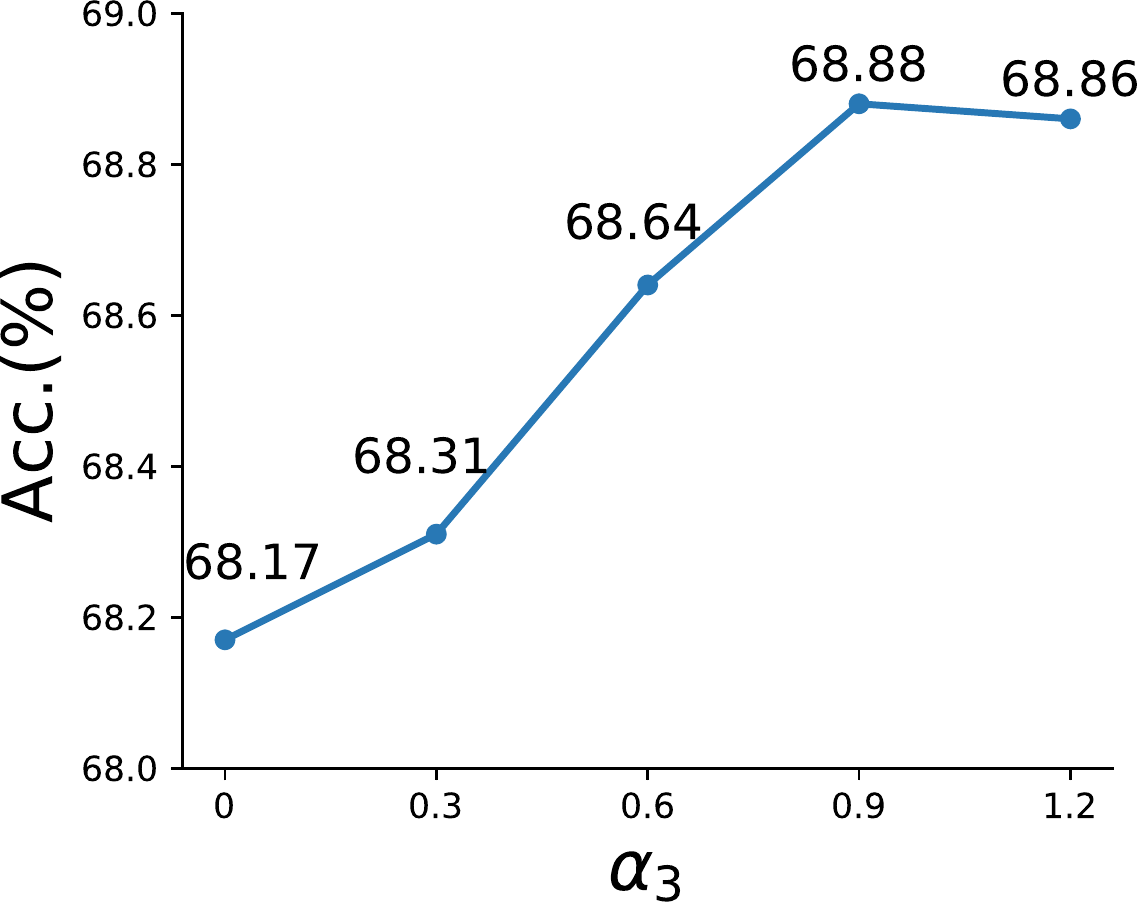} \label{hyperparameters:alpha3}
}\hspace{-3mm}
\subfigure[]{
\includegraphics[width=0.19\columnwidth]{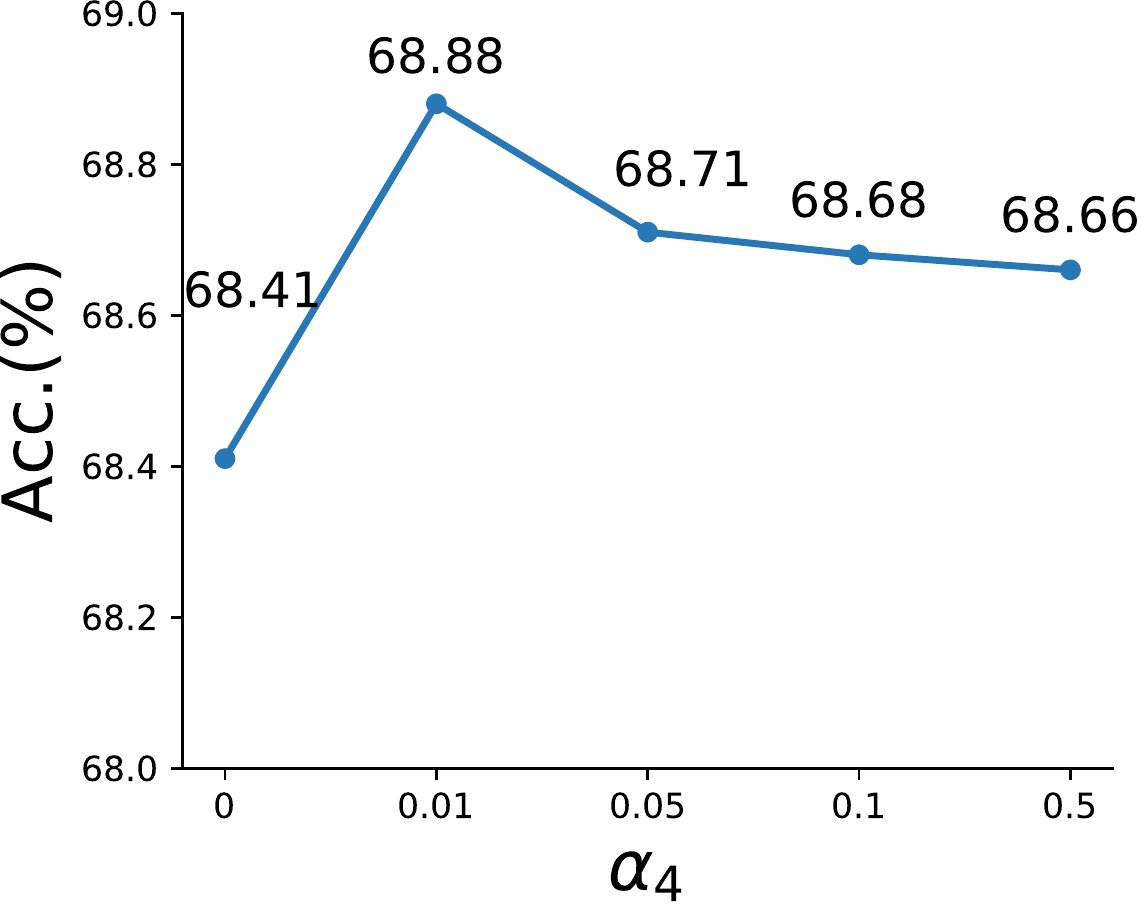} \label{hyperparameters:alpha4} 
}\hspace{-3mm}
\subfigure[]{
\includegraphics[width=0.19\columnwidth]{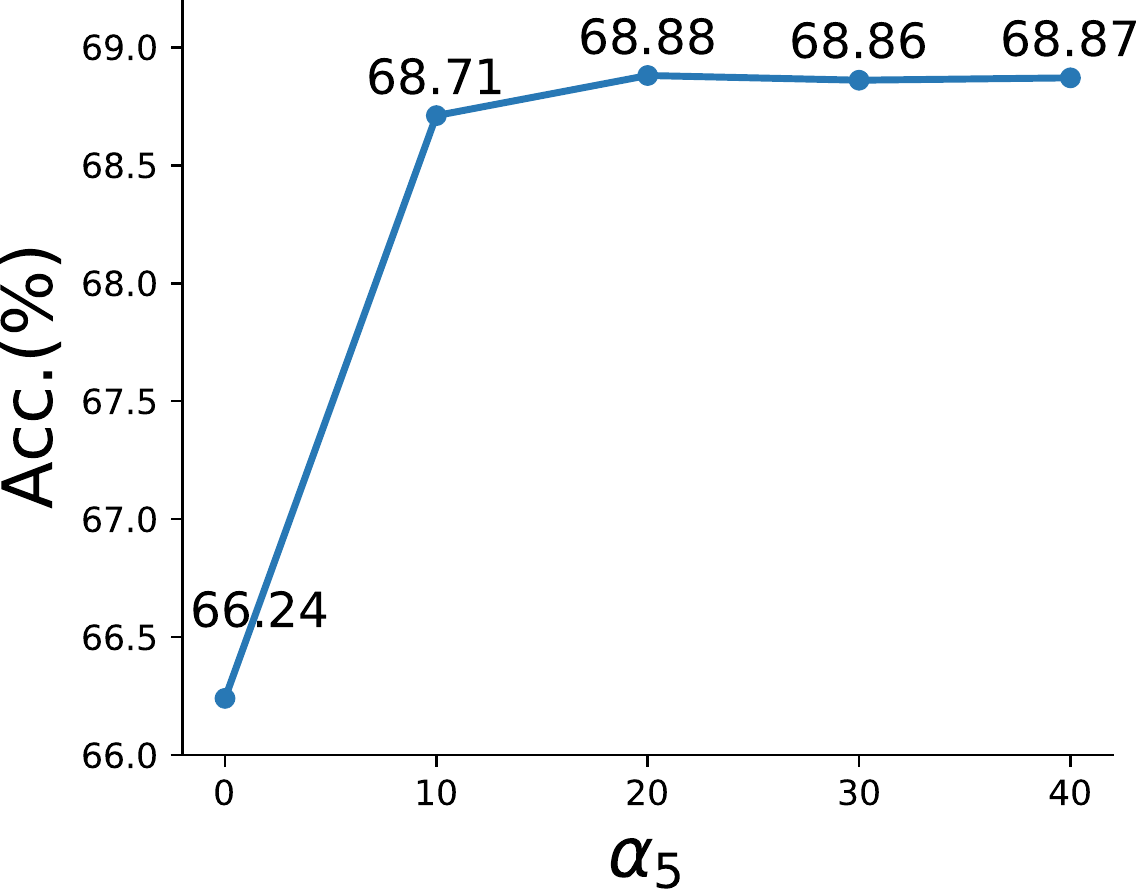} \label{hyperparameters:alpha5} 
}
\caption{Influence of the trade-off parameters.}
\label{tradeoff}
\end{figure*}

\subsection{Ablation Studies}\label{ablation}
In this section, we give an in-depth study on the influence of hyper-parameters in this paper including the trade-off parameters in Eq.\,(\ref{generator}) and Eq.\,(\ref{quantization}), and the number of available images in the calibration dataset. All experiments are conducted by quantizing all layers of ResNet-18 to 4-bit.

%
%
%
%

\subsubsection{Trade-off Parameters.}
We first display the influence of different trade-off parameters in Fig.\,\ref{tradeoff}.
The $\alpha_1$, $\alpha_2$, $\alpha_3$, and $\alpha_4$ from Eq.\,(\ref{generator}) balance different losses in updating the generator while $\alpha_5$ from Eq.\,(\ref{quantization}) balances the losses in updating the quantized model.
Each $\alpha_i$ is first empirically initialized. Then, for $\alpha_i$, we search its optimal value using the grid search with others fixed.
From Fig.\,\ref{tradeoff}, we can see that the optimal configurations of these three parameters are $\alpha_1=0.5$, $\alpha_2=0.2$, $\alpha_3 = 0.9$, $\alpha_4 = 0.05$ and $\alpha_5 = 20$, which are also our settings for all the aforementioned experiments. Though they might not be the optimal for all networks, we find these configurations already bring better performance than the recent state-of-the-arts. %
Also, we observe that $\alpha_4 << \alpha_3$, in which $\alpha_3$ and $\alpha_4$ respectively balance the importance of the proposed BNS-distorted loss and BNS-centralized loss. This is due to the distorted centroid for synthetic data is centered on the corresponding class centroid, thus the BNS-distorted loss can retain the inter-class separation to some extent, which partly relieves the involvement of BNS-centralized loss and leads to a small $\alpha_4$.

%

%

\begin{figure}[t]
\centering
\includegraphics[height=0.35\linewidth]{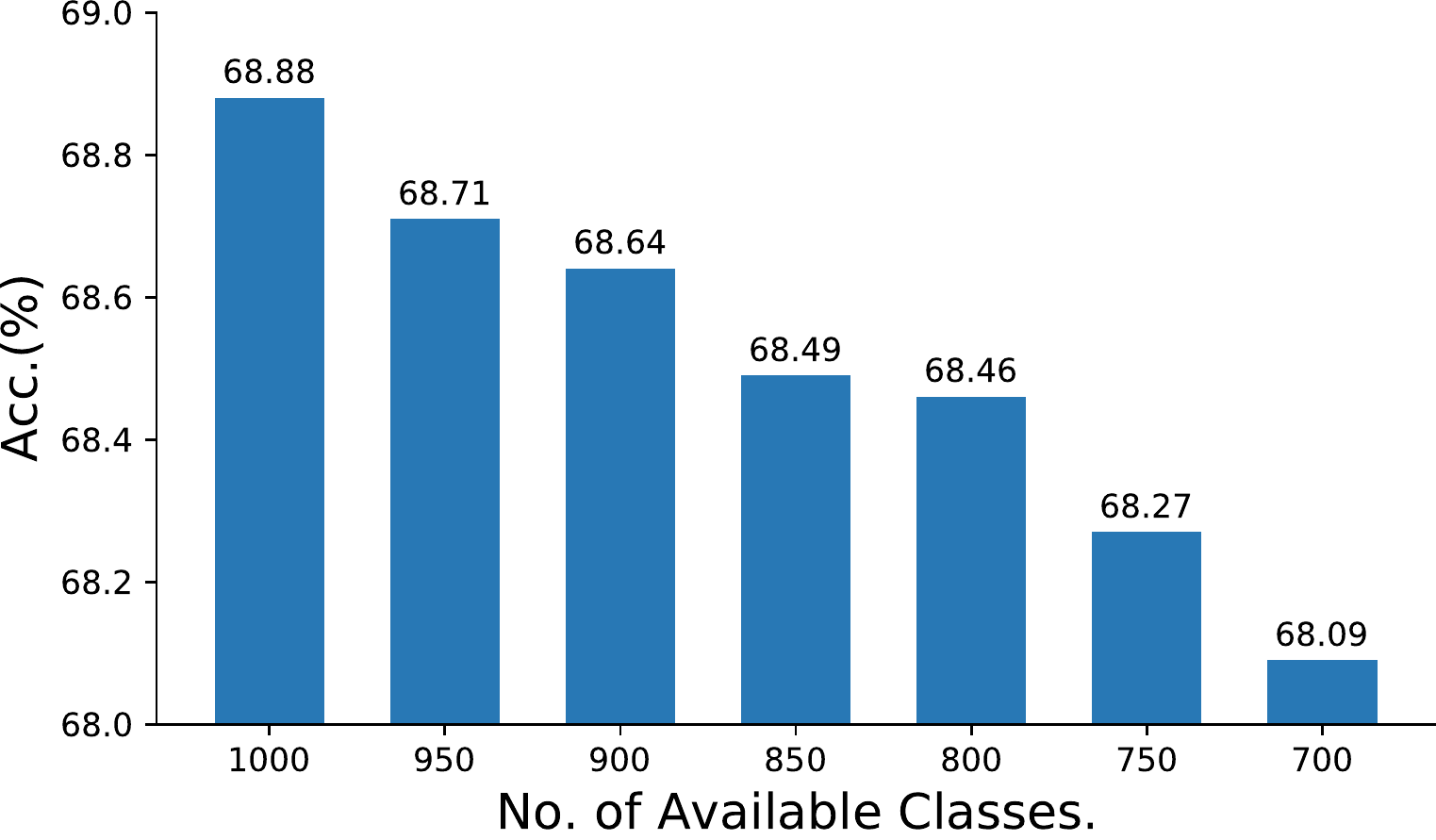}
\caption{Effect of Available Classes.}
\label{fewerclasses}
\end{figure}

\subsubsection{Effect of Available Classes.}
The calibration dataset consists of 1,000 images including one image per class by default. However, some images might be missing in real-world applications. Consequently, the class information of corresponding image is not available for our fine-grained BNS alignment.
In Fig.\,\ref{fewerclasses}, we further excavate the influence of available classes on our final performance. For unavailable classes, we omit their C-BNS and D-BNS loss when computing the Eq.\,(\ref{generator}). Note that in this case, the size of the calibration dataset is equal to the number of available classes. It can be seen from Fig.\,\ref{fewerclasses} that performance drops as the available classes decrease. Nevertheless, comparing to the recent advance, BRECQ, which obtains only 67.94\% top-1 accuracy (see Table\,\ref{comparison}), our FDDA still maintains a higher performance of 68.09\% even when only 700 classes are available. The good performance can be attributed to two reasons. On one hand, synthetic data benefits post-training quantization even though some classes are missing. On the other hand, the fine-grained data alignment helps to synthesize better images for fine-tuning the quantized model.

\section{Conclusion}
In this paper, we proposed a fine-grained data distribution alignment (FDDA) to solve the insufficient data problem in post-training quantization. We observed two important BNS properties of the inter-class separation and intra-class incohesion in the deep layers of neural network. To retain these two fine-grained distribution information, we respectively proposed the BNS-centralized loss and BNS-distorted loss. Using a real image from the calibration dataset as the centroid of each class, the BNS-centralized loss constrains the BNS of synthetic data to be close to the BNS of its class centroid, while the BNS-distorted loss introduces Gaussian noise to distort the class centroid for the purpose of incohesion. By retaining these two properties in the synthetic data, our FDDA shows its superiority over the state-of-the-art competitors on ImageNet, particularly in the hardware-friendly case where the first and last layers of networks are also quantized to low precision.

\appendix

\clearpage
%
%
\bibliographystyle{splncs04}
\bibliography{egbib}

\clearpage

\section*{Appendix \label{appendix}}

\section{More Visualization}

\subsection{Visualization of MobileNetV2}

\begin{figure*}[!h]
\centering
\subfigure[]{
\includegraphics[height=0.145\linewidth]{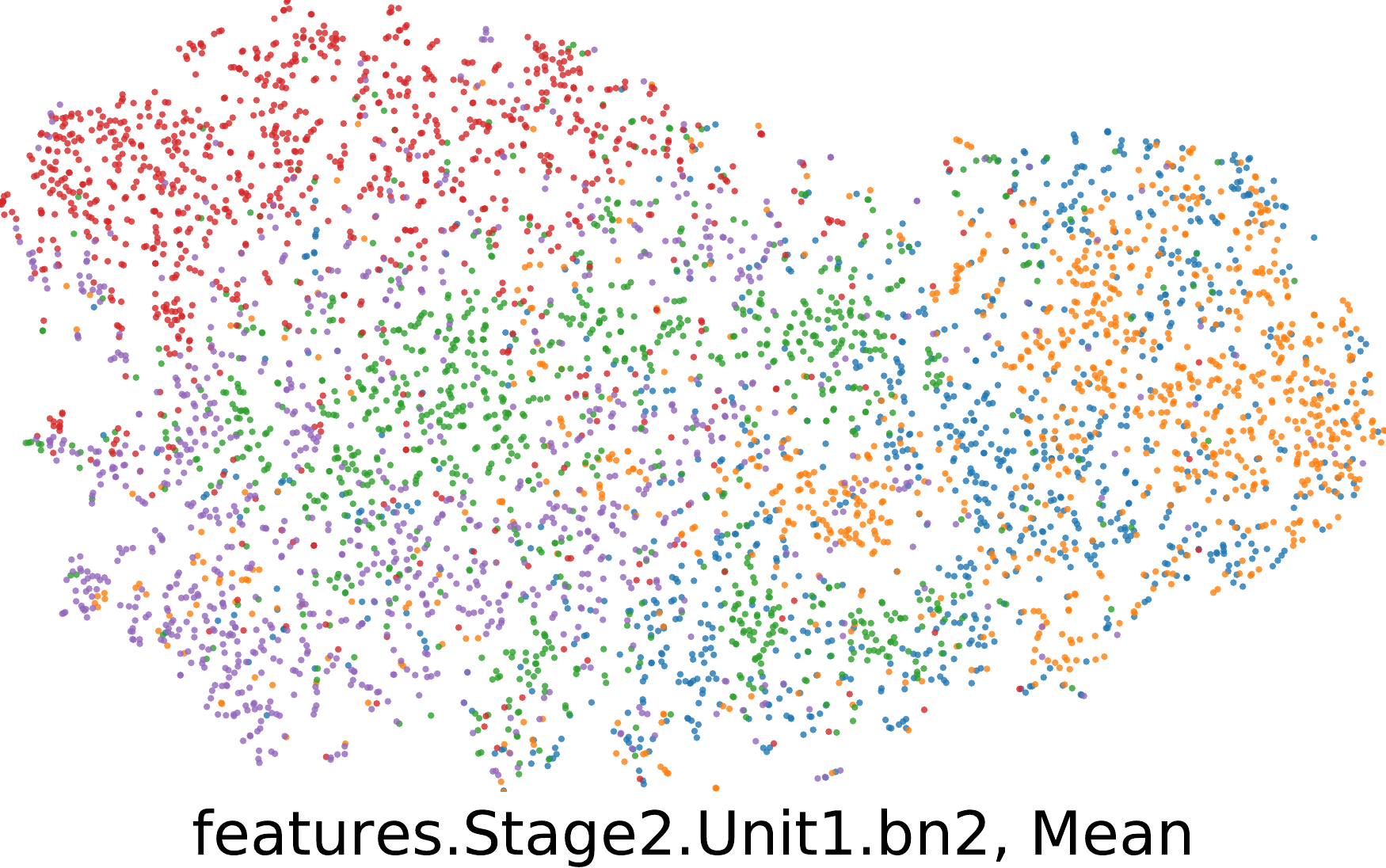} \label{visualization-mv1:1}
}
\subfigure[]{
\includegraphics[height=0.145\linewidth]{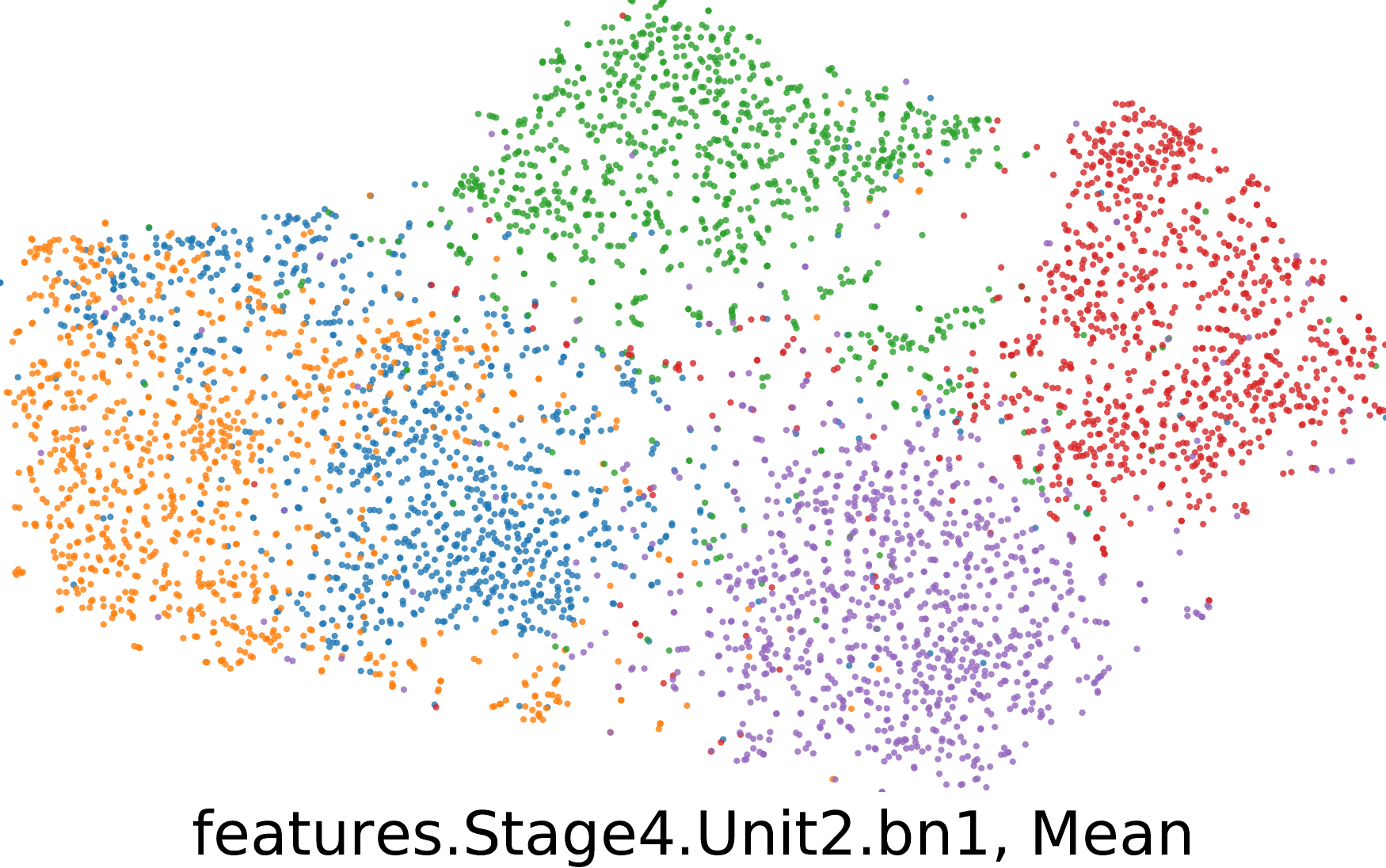} \label{visualization-mv1:2} 
}
\subfigure[]{
\includegraphics[height=0.145\linewidth]{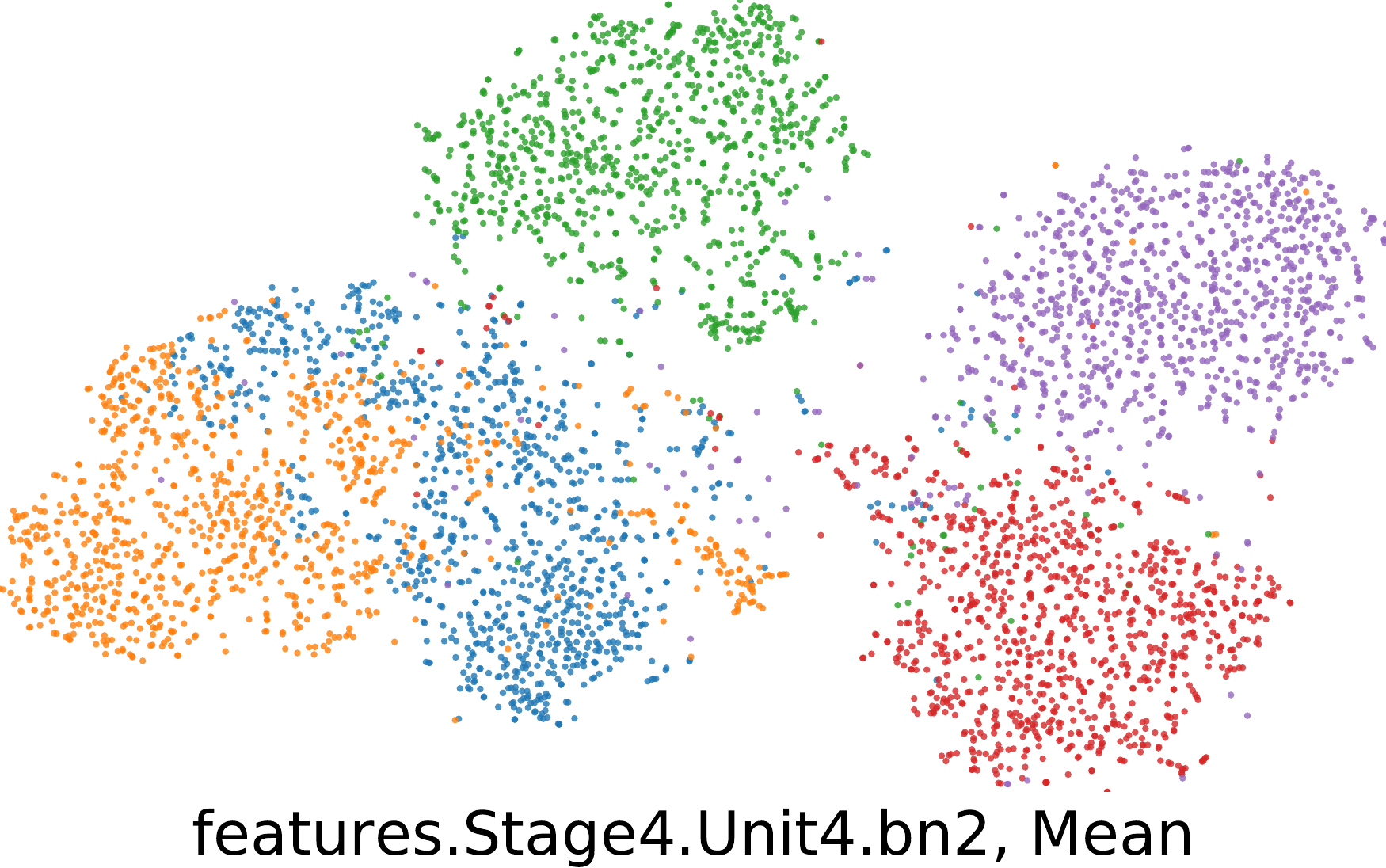} \label{visualization-mv1:3} 
}
\subfigure[]{
\includegraphics[height=0.145\linewidth]{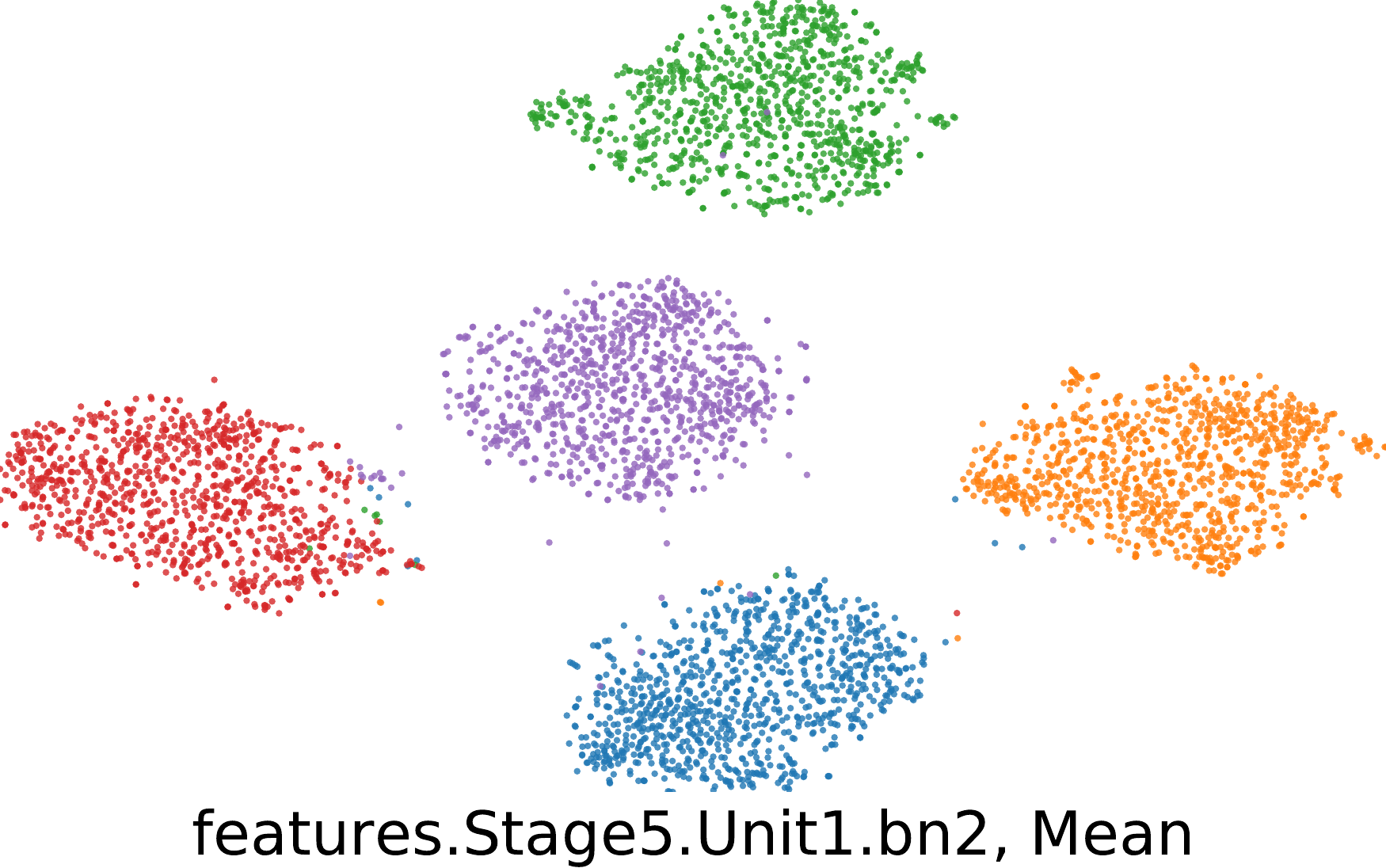}\label{visualization-mv1:4}
}

\subfigure[]{
\includegraphics[height=0.145\linewidth]{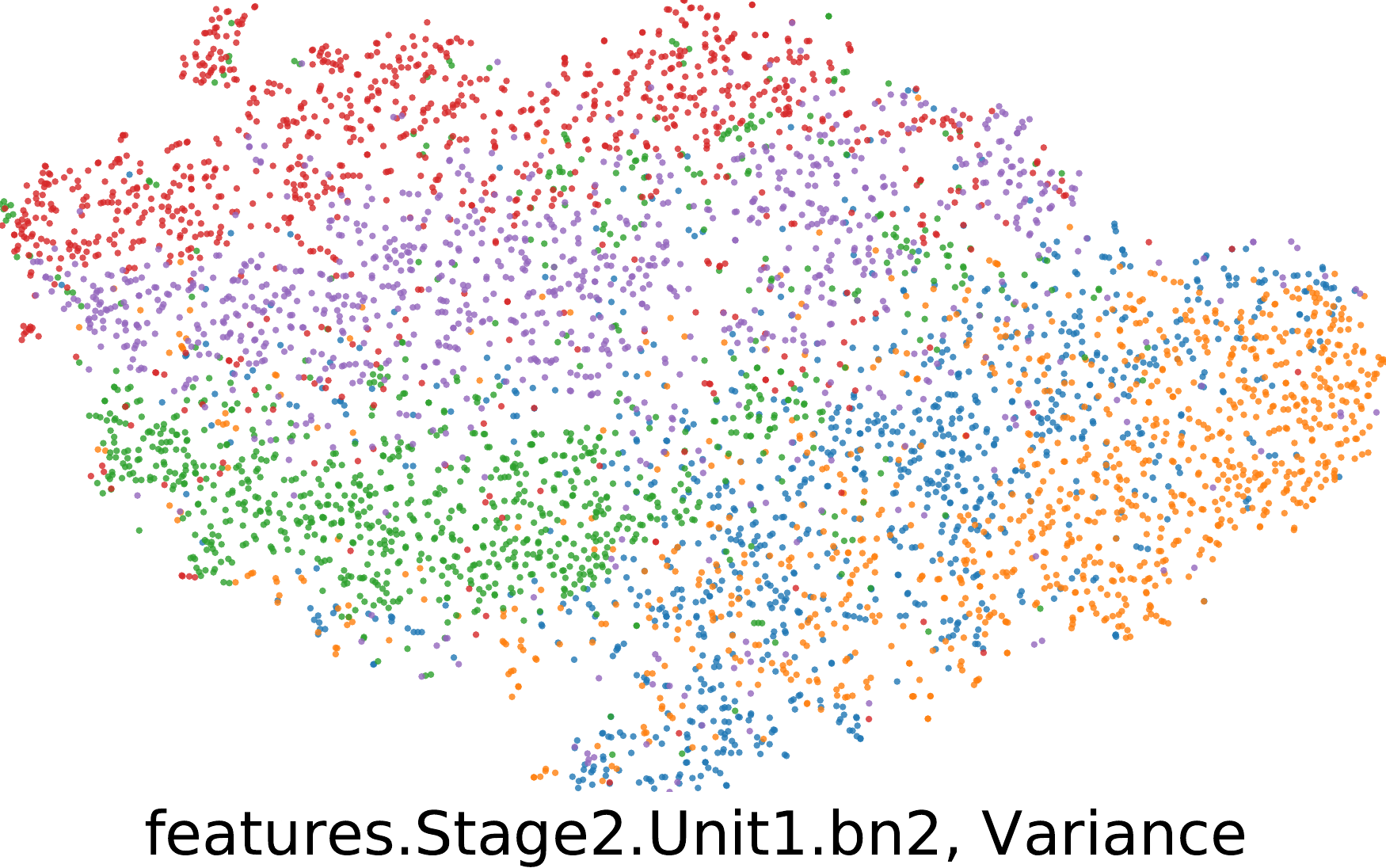} \label{visualization-mv1:5}
}
\subfigure[]{
\includegraphics[height=0.145\linewidth]{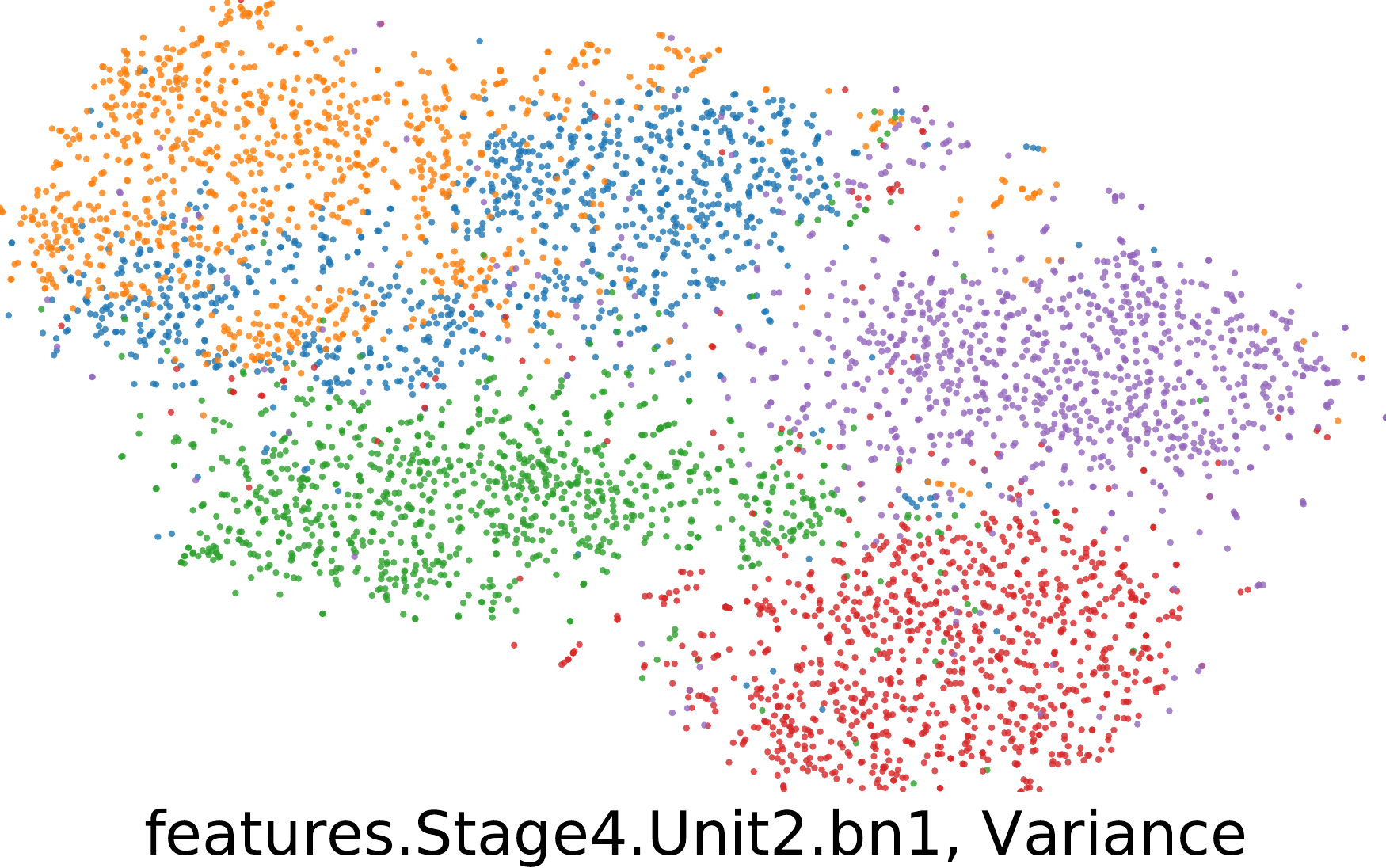} \label{visualization-mv1:6} 
}
\subfigure[]{
\includegraphics[height=0.145\linewidth]{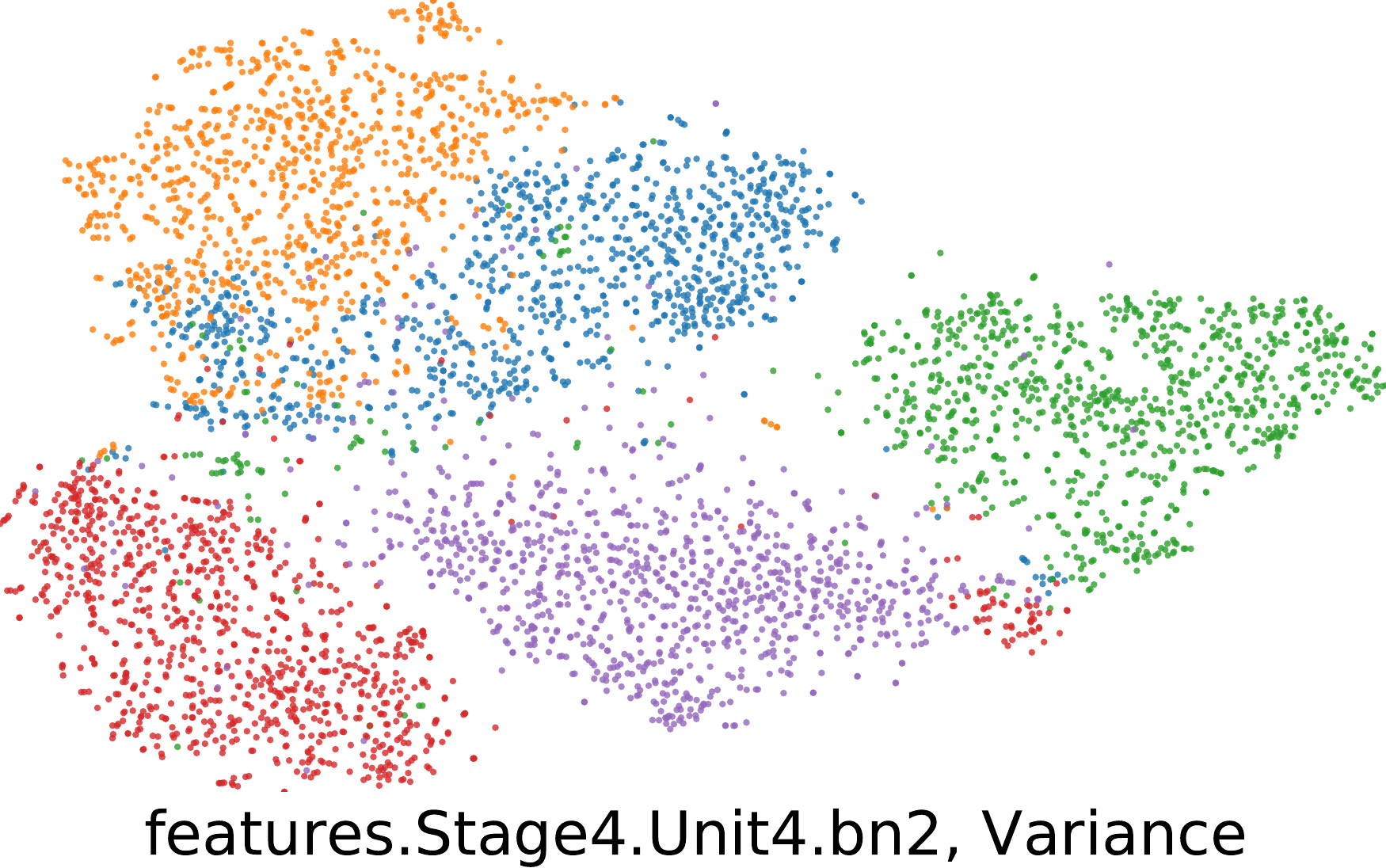} \label{visualization-mv1:7} 
}
\subfigure[]{
\includegraphics[height=0.145\linewidth]{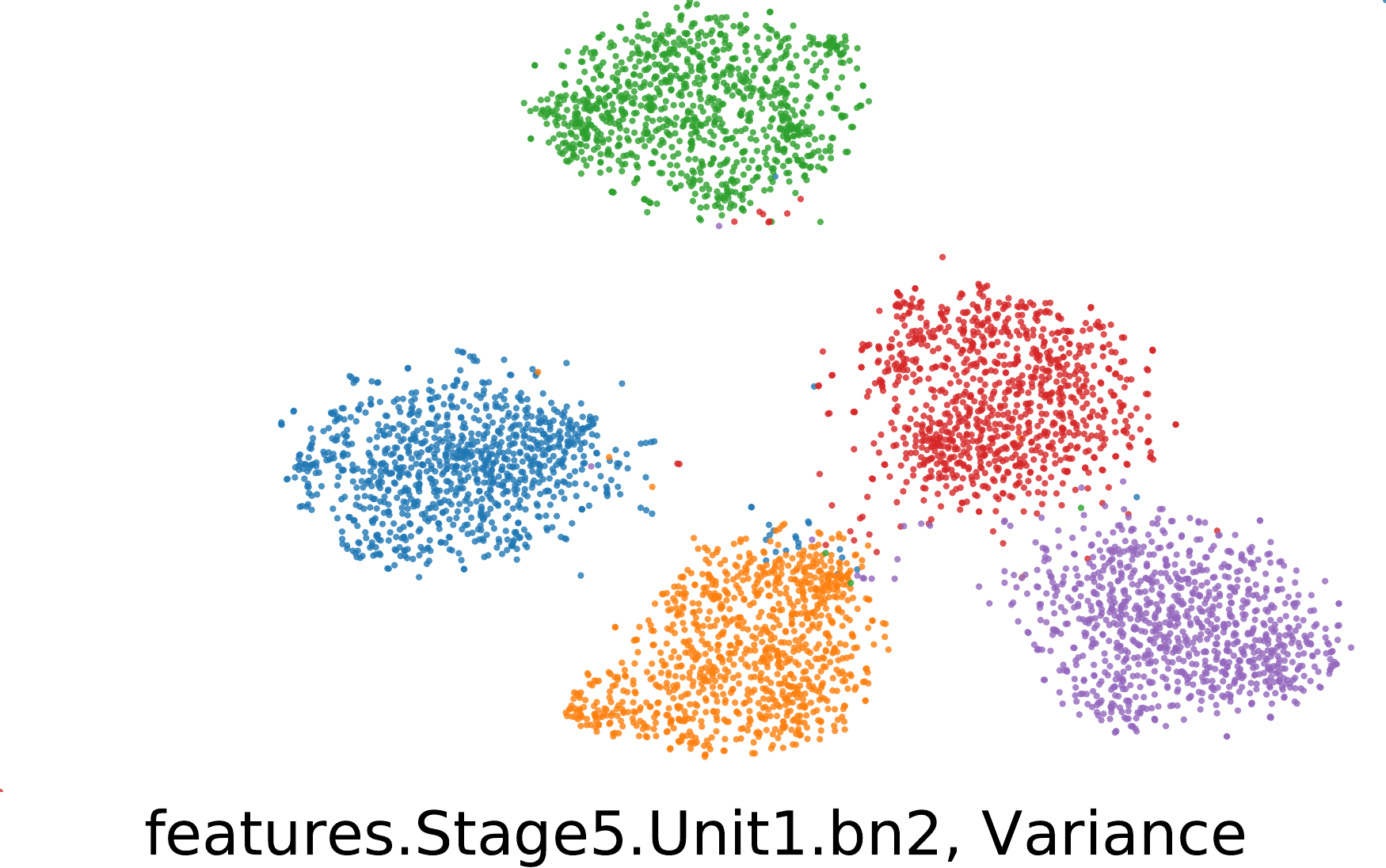}\label{visualization-mv1:8}
}
\caption{t-SNE visualization (five classes) of BNS in different layers of pre-trained MobileNetV1 on ImageNet. Best viewed in color.}
\label{visualization-mv1}
\end{figure*}

The visualization of BNS in different layers of pre-trained MobileNetV1 is shown in Fig.\,\ref{visualization-mv1}.

\clearpage

\subsection{Visualization of MobileNetV2}

\begin{figure*}[!h]
\centering
\subfigure[]{
\includegraphics[height=0.145\linewidth]{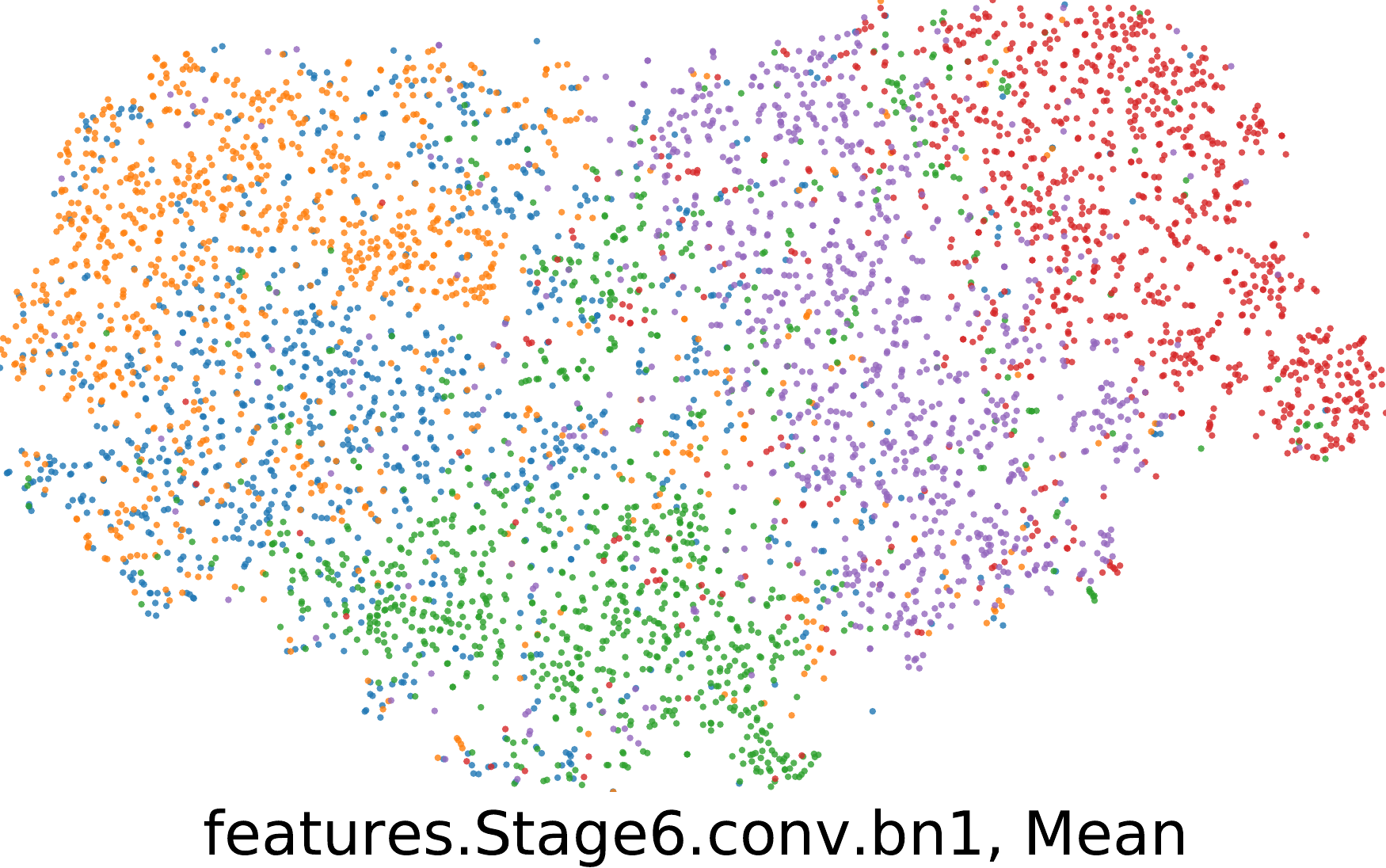} \label{visualization-mv2:1}
}
\subfigure[]{
\includegraphics[height=0.145\linewidth]{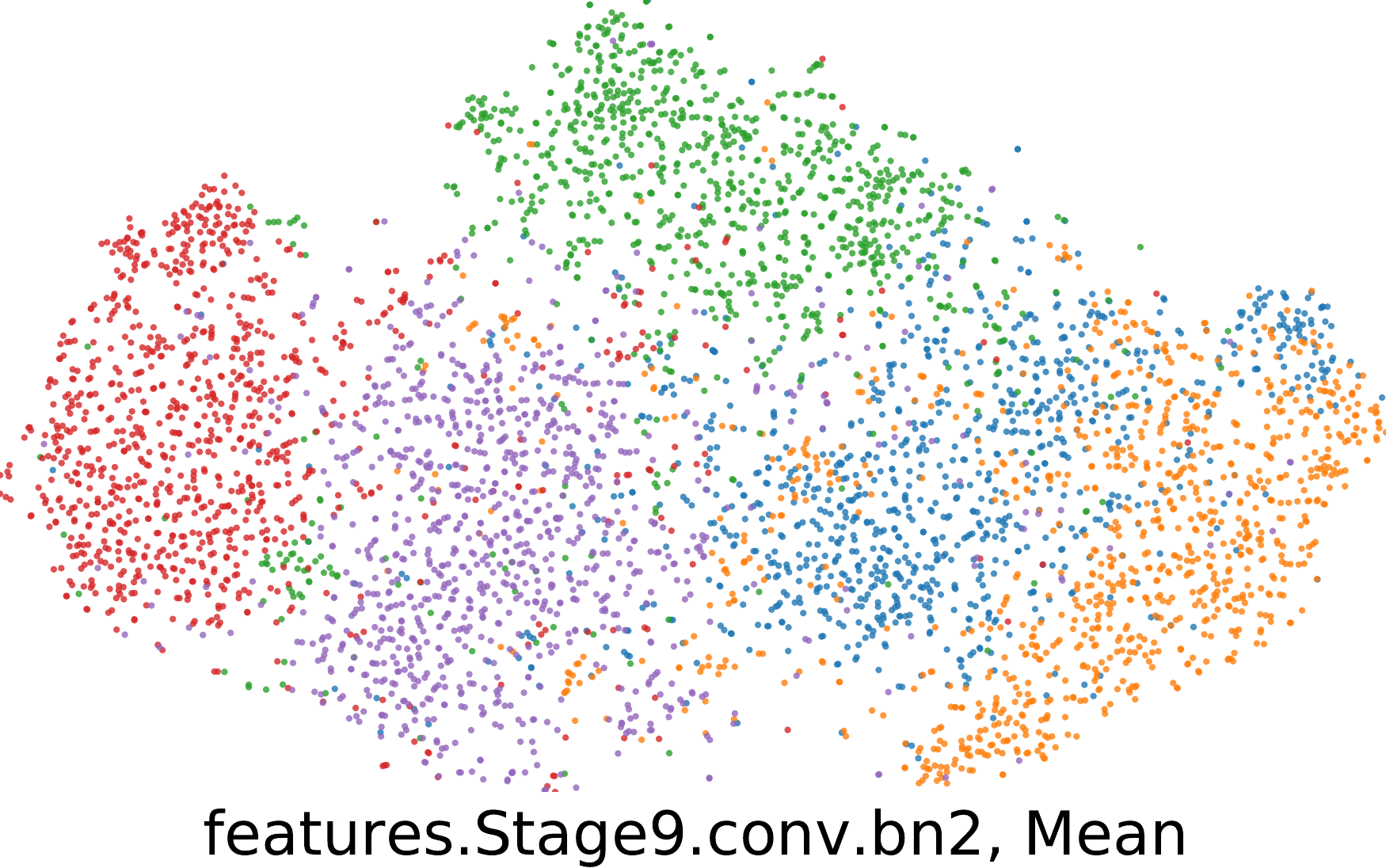} \label{visualization-mv2:2} 
}
\subfigure[]{
\includegraphics[height=0.145\linewidth]{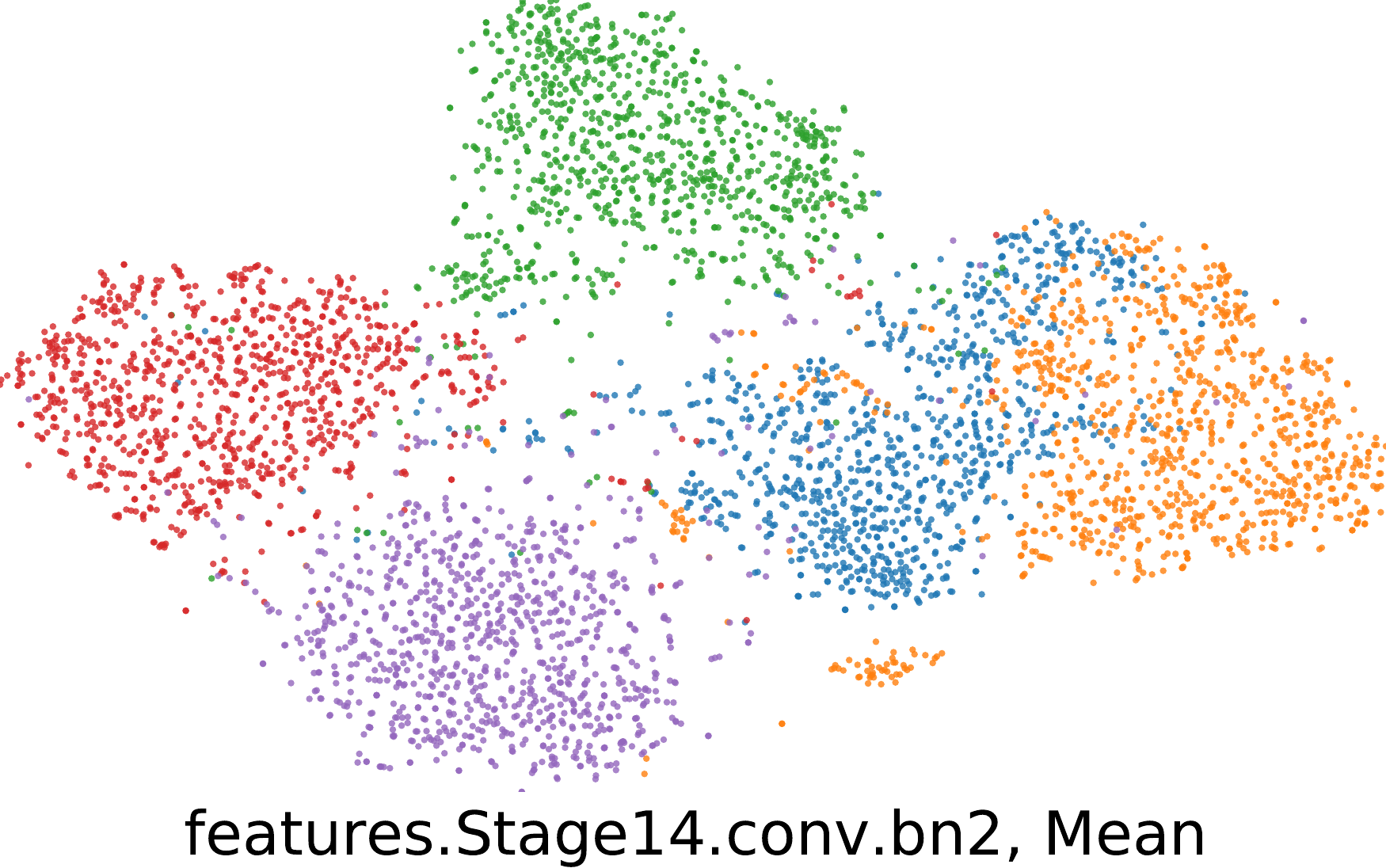} \label{visualization-mv2:3} 
}
\subfigure[]{
\includegraphics[height=0.145\linewidth]{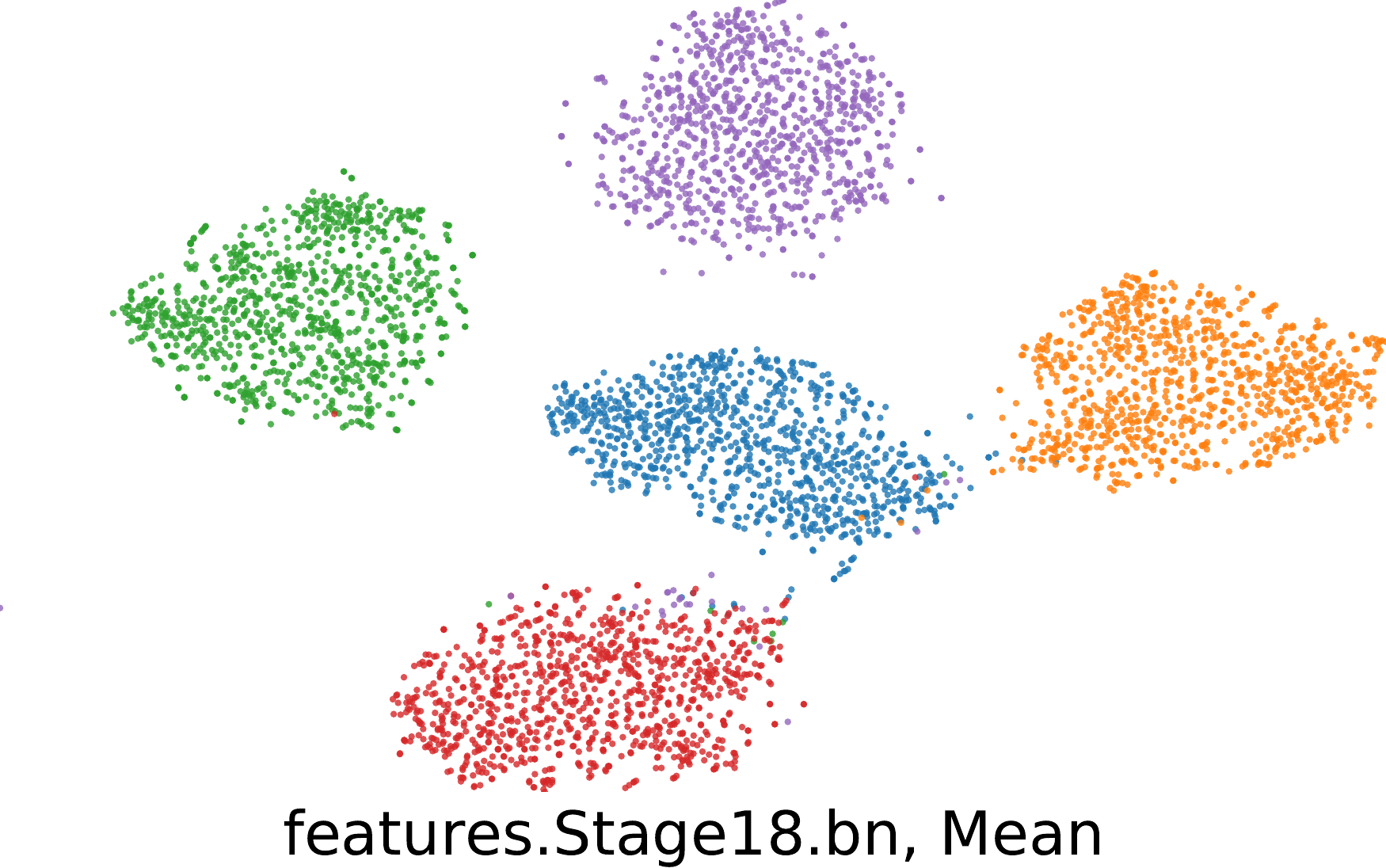}\label{visualization-mv2:4}
}

\subfigure[]{
\includegraphics[height=0.145\linewidth]{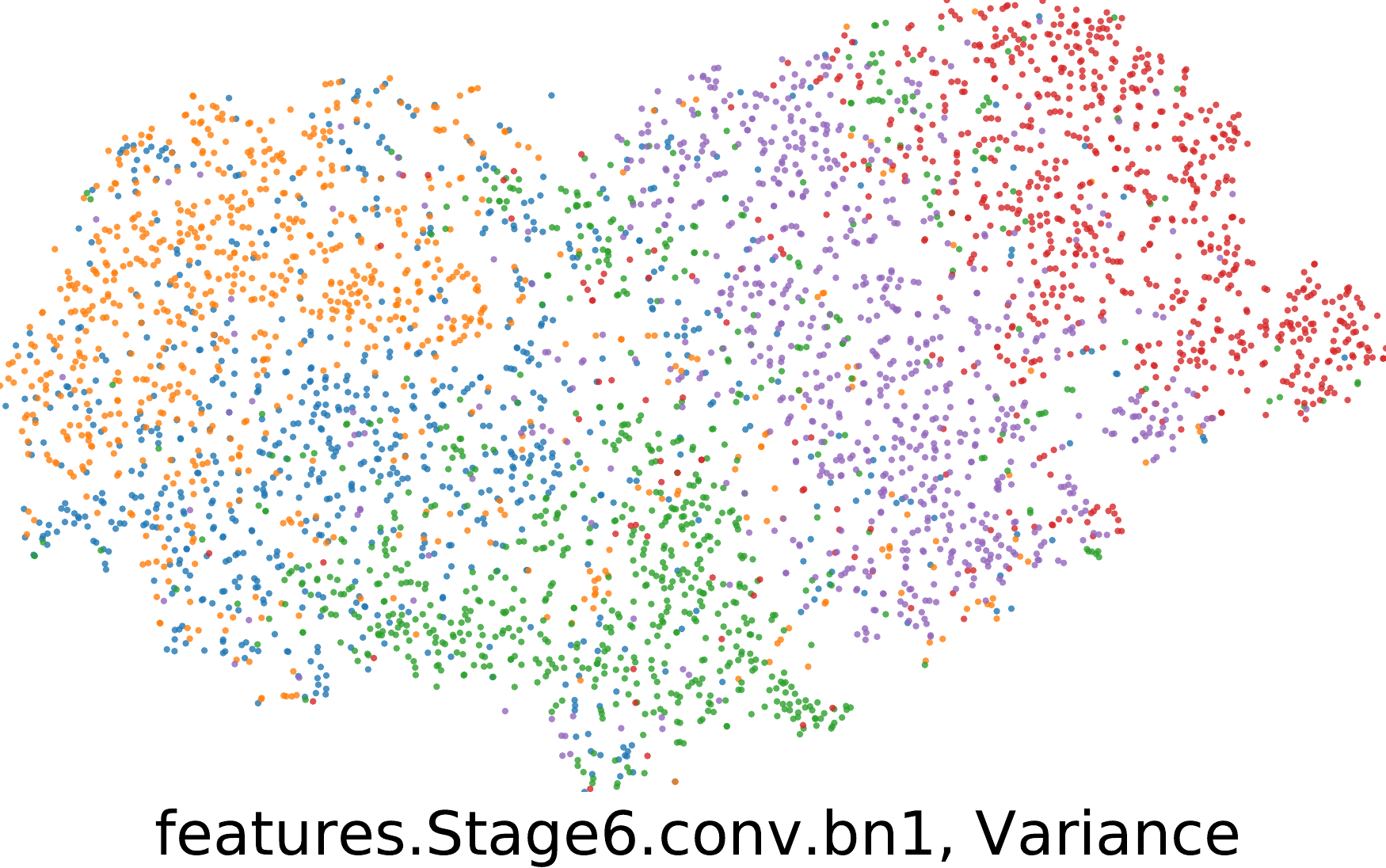} \label{visualization-mv2:5}
}
\subfigure[]{
\includegraphics[height=0.145\linewidth]{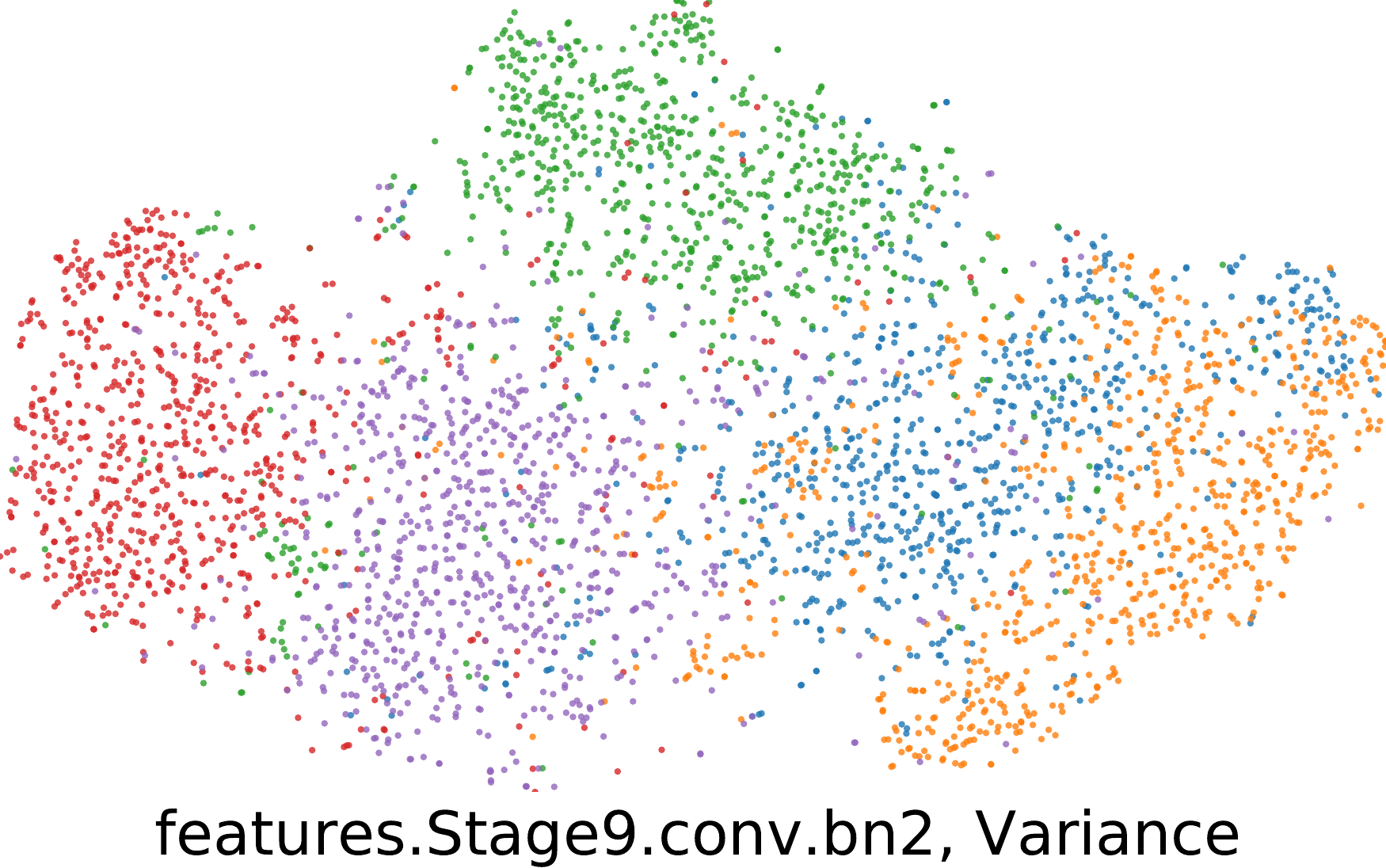} \label{visualization-mv2:6} 
}
\subfigure[]{
\includegraphics[height=0.145\linewidth]{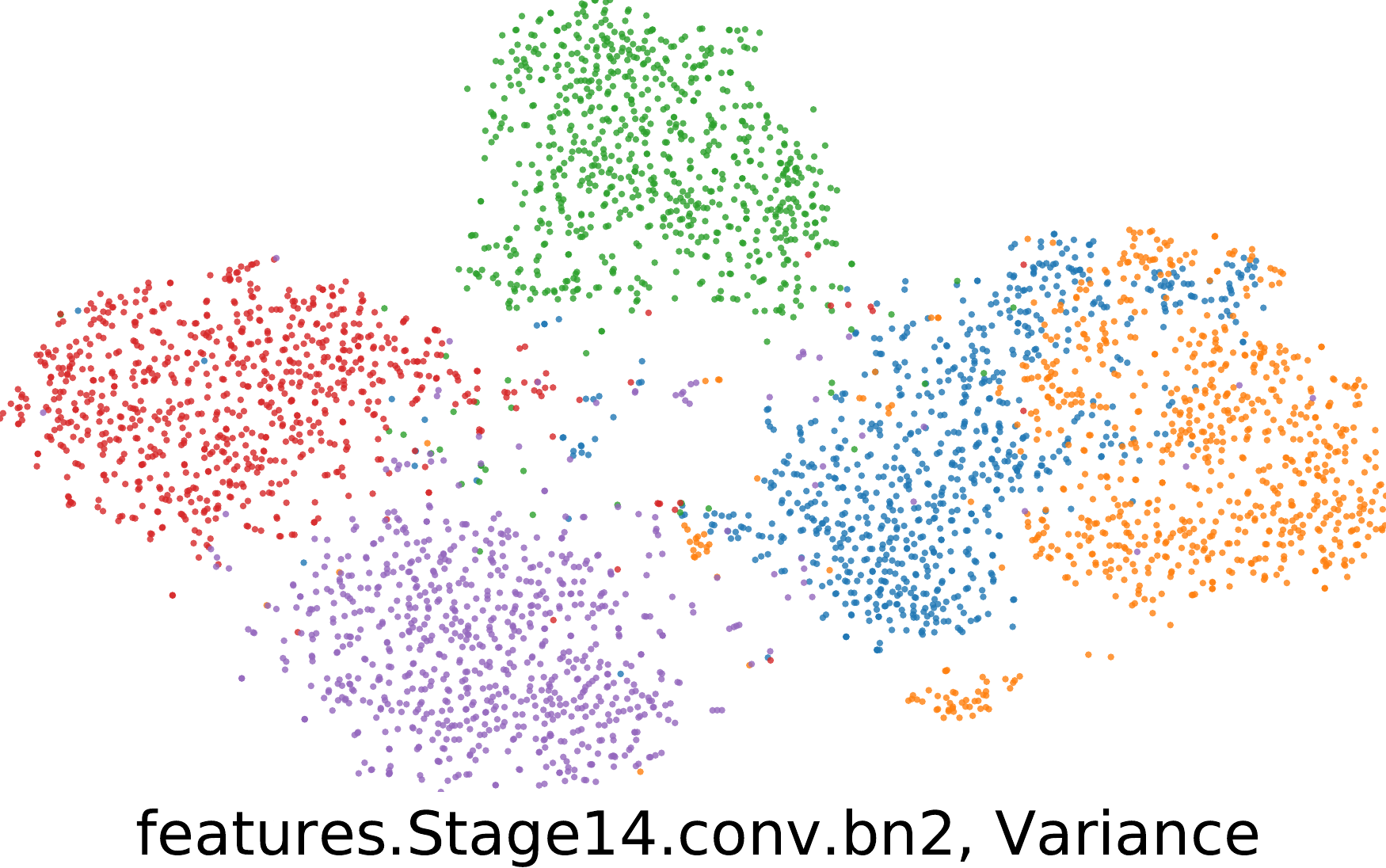} \label{visualization-mv2:7} 
}
\subfigure[]{
\includegraphics[height=0.145\linewidth]{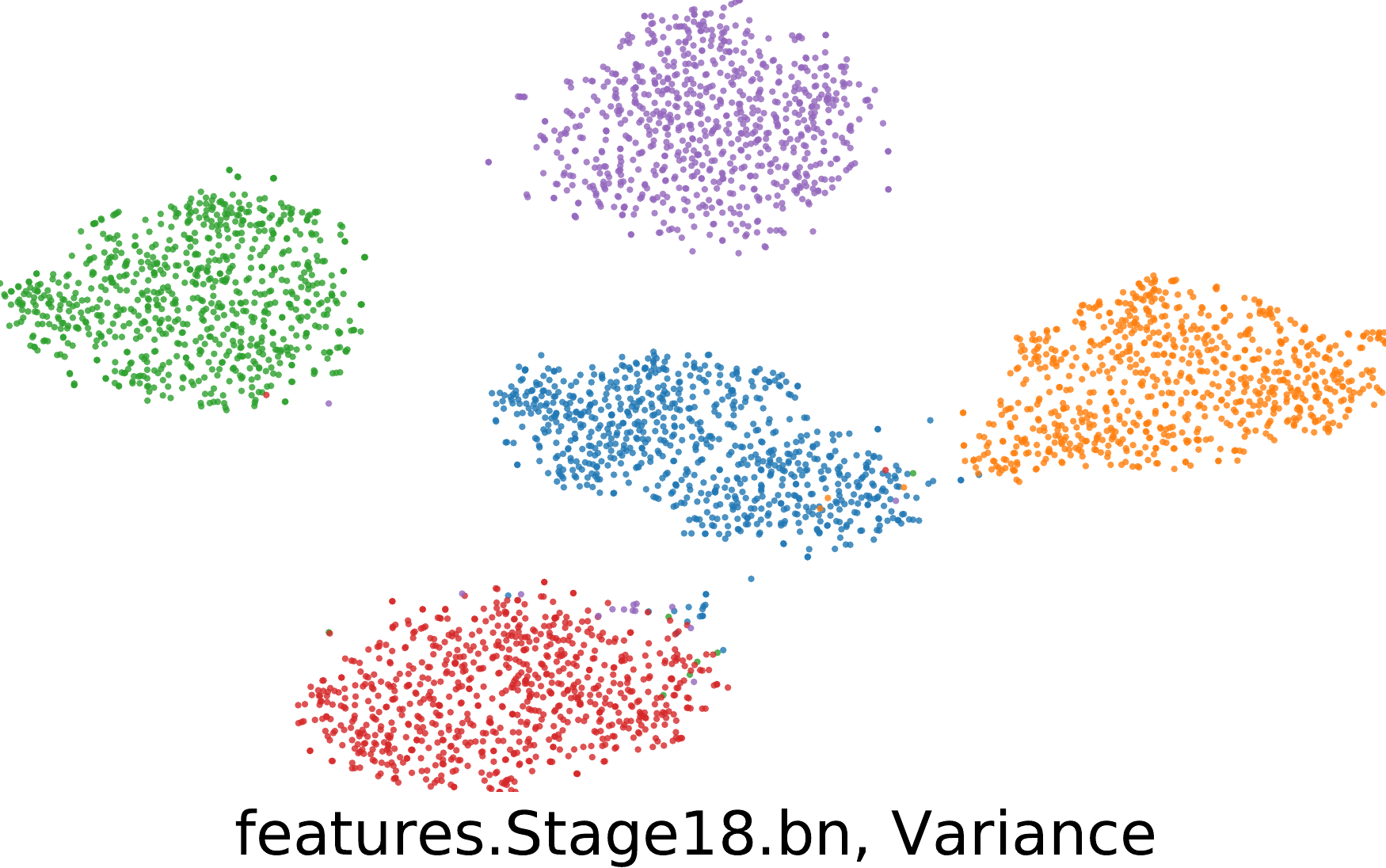}\label{visualization-mv2:8}
}
\caption{t-SNE visualization (five classes) of BNS in different layers of pre-trained MobileNetV2 on ImageNet. Best viewed in color.}
\label{visualization-mv2}
\end{figure*}

The visualization of BNS in different layers of pre-trained MobileNetV2 is shown in Fig.\,\ref{visualization-mv2}.

\end{document}


\pagestyle{headings}
\mainmatter
\def\ECCVSubNumber{190}  

\title{Supplementary for Paper 190} 

\titlerunning{ECCV-22 submission ID \ECCVSubNumber} 
\authorrunning{ECCV-22 submission ID \ECCVSubNumber} 
\author{Anonymous ECCV submission}
\institute{Paper ID \ECCVSubNumber}

\maketitle

\section{More Visualization}

\subsection{Visualization of MobileNetV2}

The visualization of BNS in different layers of pre-trained MobileNetV1 is shown in Fig.\,\ref{visualization-mv1}.

\begin{figure*}[!h]
\centering
\subfigure[]{
\includegraphics[height=0.145\linewidth]{supp-figures/Mv1-features.Stage2.Unit1.bn2, Mean.pdf} \label{visualization-mv1:1}
}
\subfigure[]{
\includegraphics[height=0.145\linewidth]{supp-figures/Mv1-features.Stage4.Unit2.bn1, Mean.pdf} \label{visualization-mv1:2} 
}
\subfigure[]{
\includegraphics[height=0.145\linewidth]{supp-figures/Mv1-features.Stage4.Unit4.bn2, Mean.pdf} \label{visualization-mv1:3} 
}
\subfigure[]{
\includegraphics[height=0.145\linewidth]{supp-figures/Mv1-features.Stage5.Unit1.bn2, Mean.pdf}\label{visualization-mv1:4}
}

\subfigure[]{
\includegraphics[height=0.145\linewidth]{supp-figures/Mv1-features.Stage2.Unit1.bn2, Variance.pdf} \label{visualization-mv1:5}
}
\subfigure[]{
\includegraphics[height=0.145\linewidth]{supp-figures/Mv1-features.Stage4.Unit2.bn1, Variance.pdf} \label{visualization-mv1:6} 
}
\subfigure[]{
\includegraphics[height=0.145\linewidth]{supp-figures/Mv1-features.Stage4.Unit4.bn2, Variance.pdf} \label{visualization-mv1:7} 
}
\subfigure[]{
\includegraphics[height=0.145\linewidth]{supp-figures/Mv1-features.Stage5.Unit1.bn2, Variance.pdf}\label{visualization-mv1:8}
}
\caption{t-SNE visualization (five classes) of BNS in different layers of pre-trained MobileNetV1 on ImageNet. Best viewed in color.}
\label{visualization-mv1}
\end{figure*}

\clearpage

\subsection{Visualization of MobileNetV2}

The visualization of BNS in different layers of pre-trained MobileNetV2 is shown in Fig.\,\ref{visualization-mv2}.

\begin{figure*}[!h]
\centering
\subfigure[]{
\includegraphics[height=0.145\linewidth]{supp-figures/Mv2-features.Stage6.conv.bn1, Mean.pdf} \label{visualization-mv2:1}
}
\subfigure[]{
\includegraphics[height=0.145\linewidth]{supp-figures/Mv2-features.Stage9.conv.bn2, Mean.pdf} \label{visualization-mv2:2} 
}
\subfigure[]{
\includegraphics[height=0.145\linewidth]{supp-figures/Mv2-features.Stage14.conv.bn2, Mean.pdf} \label{visualization-mv2:3} 
}
\subfigure[]{
\includegraphics[height=0.145\linewidth]{supp-figures/Mv2-features.Stage18.bn, Mean.pdf}\label{visualization-mv2:4}
}

\subfigure[]{
\includegraphics[height=0.145\linewidth]{supp-figures/Mv2-features.Stage6.conv.bn1, Variance.pdf} \label{visualization-mv2:5}
}
\subfigure[]{
\includegraphics[height=0.145\linewidth]{supp-figures/Mv2-features.Stage9.conv.bn2, Variance.pdf} \label{visualization-mv2:6} 
}
\subfigure[]{
\includegraphics[height=0.145\linewidth]{supp-figures/Mv2-features.Stage14.conv.bn2, Variance.pdf} \label{visualization-mv2:7} 
}
\subfigure[]{
\includegraphics[height=0.145\linewidth]{supp-figures/Mv2-features.Stage18.bn, Variance.pdf}\label{visualization-mv2:8}
}
\caption{t-SNE visualization (five classes) of BNS in different layers of pre-trained MobileNetV2 on ImageNet. Best viewed in color.}
\label{visualization-mv2}
\end{figure*}